\documentclass{article}

\usepackage{microtype}
\usepackage{graphicx}
\usepackage{subfigure}
\usepackage{booktabs} % for professional tables

\usepackage{hyperref}

\usepackage[accepted]{icml2025}

\usepackage{amsmath}
\usepackage{amssymb}
\usepackage{mathtools}
\usepackage{amsthm}

\usepackage[capitalize,noabbrev]{cleveref}

\theoremstyle{plain}

\theoremstyle{definition}

\theoremstyle{remark}

\usepackage[autostyle]{csquotes}
\usepackage{bm} % bm
\usepackage{amsfonts}
\usepackage{hyperref} 
\usepackage{xcolor} % colors
\usepackage{natbib}
\definecolor{mydarkblue}{rgb}{0,0.08,0.45}
\hypersetup{
colorlinks=true,
linkcolor=mydarkblue,
citecolor=mydarkblue,
filecolor=mydarkblue,
urlcolor=mydarkblue}

\DeclareMathOperator*{\argmin}{argmin}

\newcommand{\astfootnote}[1]{%
\let\oldthefootnote=\thefootnote%
\setcounter{footnote}{0}%
\renewcommand{\thefootnote}{\fnsymbol{footnote}}%
\footnote{#1}%
\let\thefootnote=\oldthefootnote%
}

\def\rvh{{\mathbf{h}}}

\def\rvr{{\mathbf{r}}}

\def\rvx{{\mathbf{x}}}

\def\rvz{{\mathbf{z}}}

\def\rmA{{\mathbf{A}}}

\def\rmH{{\mathbf{H}}}

\def\rmK{{\mathbf{K}}}

\def\rmP{{\mathbf{P}}}

\def\rmR{{\mathbf{R}}}

\def\rmV{{\mathbf{V}}}
\def\rmW{{\mathbf{W}}}

\DeclareMathAlphabet{\mathsfit}{\encodingdefault}{\sfdefault}{m}{sl}
\SetMathAlphabet{\mathsfit}{bold}{\encodingdefault}{\sfdefault}{bx}{n}

\usepackage{xcolor} % colors
\usepackage{pdflscape}

\newif\ifshowcomments
\showcommentstrue % Uncomment this line to show comments
\newif\ifcameraready
\camerareadytrue % Uncomment for camera ready. Submission should be anonymous
\camerareadyfalse % Uncomment for anonymous submission version

\icmltitlerunning{Neural Augmented Kalman Filters for Road Network assisted GNSS positioning}

\begin{document}

\twocolumn[
\icmltitle{Neural Augmented Kalman Filters for \\ Road Network assisted GNSS positioning}



\icmlsetsymbol{equal}{*}

\begin{icmlauthorlist}
\icmlauthor{Hans van Gorp}{equal,uni,qc}
\icmlauthor{Davide Belli}{equal,qc}
\icmlauthor{Amir Jalalirad}{qc}
\icmlauthor{Bence Major}{qc}
\end{icmlauthorlist}

\icmlaffiliation{uni}{Department of Electrical Engineering, Eindhoven University of Technology}
\icmlaffiliation{qc}{Qualcomm AI Research. Work done during an internship at Qualcomm AI Research, Amsterdam. Qualcomm AI Research is an initiative of Qualcomm Technologies, Inc.}

\icmlcorrespondingauthor{Hans van Gorp}{h.v.gorp@tue.nl}
\icmlcorrespondingauthor{Davide Belli}{dbelli@qti.qualcomm.com}

\icmlkeywords{gnss, positioning, road networks, kalman filter, viterbi, graph, neural networks, lstm}

\vskip 0.3in
]



\printAffiliationsAndNotice{\icmlEqualContribution} % otherwise use the standard text.

% \ifcameraready
% \footnotetext[2]{Qualcomm AI Research is an initiative of Qualcomm Technologies, Inc.}
% \fi

\begin{abstract}
%The Global Navigation Satellite System (GNSS) provides critical positioning and timing information globally, but its accuracy in dense urban environments is often compromised by multipath and non-line-of-sight errors. This paper introduces a novel approach to enhance GNSS-based vehicle positioning by integrating road network information into a Kalman Filter (KF) using a Temporal Graph Neural Network (TGNN). The TGNN is designed to predict the correct road segment and its associated uncertainty, improving the measurement update step of the KF.  We validate our approach with real-world GNSS data and open-source road network information. Our TGNN outperforms traditional Viterbi-based approaches, reducing the horizontal positioning error by 29\% in challenging urban scenarios.
The Global Navigation Satellite System (GNSS) provides critical positioning information globally, but its accuracy in dense urban environments is often compromised by multipath and non-line-of-sight errors. Road network data can be used to reduce the impact of these errors and enhance the accuracy of a positioning system. Previous works employing road network data are either limited to offline applications, or rely on Kalman Filter (KF) heuristics with little flexibility and robustness. We instead propose training a Temporal Graph Neural Network (TGNN) to integrate road network information into a KF. The TGNN is designed to predict the correct road segment and its associated uncertainty to be used in the measurement update step of the KF. We validate our approach with real-world GNSS data and open-source road networks, observing a 29\% decrease in positioning error for challenging scenarios compared to a GNSS-only KF. To the best of our knowledge, ours is the first deep learning-based approach jointly employing road network data and GNSS measurements to determine the user position on Earth.
\end{abstract}
\section{Introduction}

\begin{figure}[tbh!]
    \centering
    \includegraphics[width=0.99\linewidth]{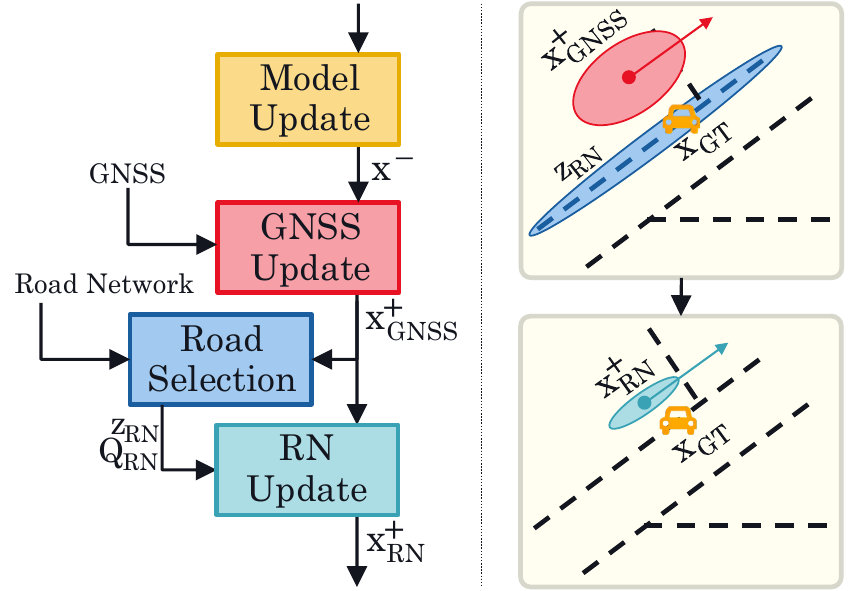}
    \caption{In order to leverage road network data in a GNSS-based KF, a Gaussian observation (mean $z$, covariance $Q$) needs to be constructed by selecting the correct road segment. In our proposed method, this road selection is performed by a Temporal Graph Convolutional Network which also predicts the appropriate covariance (left). The position and uncertainty predicted by the KF can be improved using this additional data source (right).} 

    \vspace{-0.5em}
    \label{fig:figure_1}
\end{figure}

The Global Navigation Satellite System (GNSS) provides accurate position and time information to billions of receivers worldwide. In 2023, an estimated 5.6 billion GNSS receivers have been produced, with this number expected to grow to 9 billion by 2033 \citep{report}. Of these, 10\% were used in the road and automotive segment in 2023, growing to 15\% by 2033. The coverage of GNSS is extensive, with around 120 satellites active in multiple constellations (GPS, Galileo, GLONASS, BeiDou), enabling localization 
% (\hg{UK: localiSation, US: localiZation. Are we going for UK or US spelling in the manuscript?} \db{I would always stick with US spelling. I have never read 'summarisation' or 'normalisation', which sound very weird to me})
systems anywhere on Earth. 

% However, even with this high number of satellites, accurate positioning in dense urban environments remains a challenge. Tall buildings can reflect the signal, leading to multipath errors, or block the direct view of the satellite, leading to non-line-of-sight errors. 
However, even with this high number of satellites, accurate positioning in dense urban environments remains a challenge. Tall buildings can reflect the signal, leading to multipath errors, or block the direct view of the satellite, leading to non-line-of-sight errors. This is particularly problematic for applications such as lane detection and side-of-the-street detection for car navigation, delivery, and taxi services, which typically require positioning accuracies of 2 to 10 meters. However, positioning services relying solely on GNSS can result in errors up to 20 meters in urban canyons \citep{zhong2023optimizing}. Even with complex 3D models, errors of no less than 10 meters are observed \citep{zhong2023optimizing}. Leading companies use 3D Model Assisted solutions \citep{van2021end} to improve positioning in urban environments, underscoring the limitations of GNSS-only localization in challenging scenarios. Addressing these challenges is crucial for enhancing the accuracy and reliability of positioning systems in urban settings.

% \db{Consider changing the order of the following paragraphs to better lead to our contribution (we can discuss offline, just suggesting here). My proposal is: 1) Introduce type of data used for positioning. 2) mention KFs and PFs as solution, and describe how PF have strengths but also limitations due to which they are rarely used in practice. 3) Describe how using road network in KF systems is not trivial, and what previous work do. 4) Describe how we solve this problem, with our list of contributions.}

In automotive applications, positioning errors can be mitigated by introducing additional sources of information, such as vehicle motion sensors, cameras, radar and lidar \citep{skog2009survey}. All of these require the integration of additional sensors besides the GNSS receiver, which are expensive or can be unavailable. In our work, we will instead focus on the integration of road network data, a cost-effective alternative which can be accessed globally from open-source platforms \citep{OpenStreetMap}.

Kalman Filters (KFs) are the standard for processing GNSS measurements due to their efficiency, ability to fuse multiple sources of information, and the relatively affordable assumption of uni-modal Gaussian distributions for state and observations.
% Models such as Kalman Filters (KFs) and Particle Filters can be employed to jointly track the vehicle dynamics and integrate the noisy observations from multiple sensors. Particle Filters are expressive but computationally expensive, while Kalman Filters are efficient but assume uni-modal Gaussian distributions for state and observations. The uni-modal assumption is usually acceptable for GNSS and IMU sensors, thus KFs are often preferred for consumer-grade navigation systems. 
While assuming normal distributions is reasonable for GNSS measurements, the same does not hold for road network observations, which makes their integration in a KF not straightforward. If one constructs a Gaussian observation with respect to each nearby road, the total observation becomes a multi-modal Gaussian. To address this, previous solutions first select a single \enquote{best} road segment and then construct a Gaussian observation around it. The selection can be based on belief theory \citep{el2005road} or on Hidden Markov Models (HMM) combined with a Viterbi decoder \citep{atia2017lane}, the latter of which has also been used in map-matching techniques \citep{hu2023amm,song2023mapmatching,zhong2024enhancing}. 

% Alternatively, Particle Filters (PF) that do not restrict the observation to be Gaussian can be leveraged in conjunction with road network information \citep{davidson2010pf,kempinska2016pf}. However, the number of particles that need to be tracked can be very large, imposing a significant computational burden on the system. Additionally, PFs can suffer from a low number of effective particles (where many particles are used to track the same state) or even degeneracy (where only one particle is effectively being utilized).

% \db{not strictly related to the road network problem we are tackling, so would remove this paragraph}, Another solution is the generation of a virtual fish-eye sky image leveraging Google Earth data \citep{suzuki2015fisheye}. By generating a fish-eye image of the sky from the user's perspective, it can be accurately calculated which satellites are occluded by tall buildings, and thus their signals should be discarded when calculating the user's position. 

Improving over previous approaches, we propose to augment the KF with a lightweight Temporal Graph Neural Network to select the road segment for the measurement update step as well as its uncertainty (see Figure \ref{fig:figure_1}). % We show that such a GNN can predict the correct segment more accurately than the classical Viterbi decoder, and reduces the KF horizontal positioning error by 29\%. 
The main contributions in this paper are three-fold:

\begin{enumerate}
    % \vspace{-0.2em}
    \item To the best of our knowledge, we are the first to propose a deep learning method to introduce road network inputs in a GNSS-based KF. We implement this as a Temporal Graph Neural Network predicting which road segments best match the GNSS trajectory from a moving vehicle. The predicted segments are used as additional measurements to update the KF state.
    % \vspace{-0.2em}
    \item We introduce a standard deviation prediction head to regulate the impact of the road network observation in the KF update. We optimize this component end-to-end to minimize the positioning error.
    % \vspace{-0.2em}
    \item We evaluate the proposed method on challenging real-world GNSS measurements paired with open-source road network data, showing improvements over both a GNSS-only KF and a KF + Viterbi algorithm. We also include detailed ablations and qualitative analyses to assess the contribution from each component and the behavior in different scenarios. %[Alternatively: We compare the proposed method against standard KF as well as KF integrated with causal Viterbi, showing consistent improvements on a dataset consisting of real-world GNSS measurements paired with open-source road network data.]
\end{enumerate}

% \begin{itemize}
%     \vspace{-0.5em}
%     \item We propose a GNN that causally selects the road segment using the current KF state, trained on data created using a Viterbi decoder on the entire trajectory (non-causal).
%     \vspace{-0.5em}
%     \item We show that this GNN can also be trained to output the optimal standard deviation for the observation that is put around the selected road segment in the KF.
%     \vspace{-0.5em}
%     \item We compare the method to the standard KF as well as a KF integrated with a causal Viterbi algorithm for road network observations.
% \end{itemize}

% The rest of this paper is structured as follows. In the methods section, we will first explain the Viterbi algorithm used for road selection, which was subsequently utilized with the ground truth locations to find the ground truth road segments as well as to create a strong causal baseline utilizing the KF locations. We will then briefly introduce the KF, after which we will explain the GNN design used to both select the current road segment as well as the correct standard deviation to be used in the KF. In the results section, both qualitative and quantitative results are shown for the proposed method, as well as the baselines and several ablations. The paper will conclude with a discussion section. \db{Given the 7 pages limit, would avoid this paragraph, and just have an intro sentence ad the begining of each section to briefly describe its content, if needed.}

\section{Related Work}
\label{sec:related}

\paragraph{GNSS positioning with Neural Networks}
Recent work has extensively explored the use of Deep Learning methods to improve GNSS-based positioning systems. Some approaches process GNSS measurements with MLPs \citep{suzuki2021nlos} or Neural Radiance Fields \citep{neamati2023neural} to predict which measurements are in line of sight. Other methods learn to estimate the pseudo-range error through Graph Neural Networks \citep{ml2} or Long Short-Term Memory Networks \citep{zhang2021prediction}. Neural Networks can also be trained to directly produce location corrections with respect to an input anchor point \citep{kanhere2022improving,siemuri2021improving}. 

\paragraph{Neural Augmentation of Kalman Filters}
While previously mentioned works tackled the instantaneous positioning scenario, others investigated deep learning methods to track the receiver position over time. GNSS tracking solutions often consist of classical systems such as KFs, in which some of the components are augmented with data-driven Neural Networks \citep{shlezinger2024ai}.
\citet{revach2022kalmannet} proposes learning the optimal Kalman gain through the use of Recurrent Neural Networks (RNN), and \citet{li2023deep} combines CNNs and LSTMs for similar purposes. \citet{mohanty2024tightly} uses GNNs to estimate the state and state uncertainty matrices in the KFs, while \citet{gao2020rl} and \citet{han2021precise} apply Reinforcement Learning to learn the process noise covariance matrix. 
KFs can also be augmented with Neural Networks to process measurements from multiple sensors such as GNSS and Inertial Measuring Units (IMU) \citep{guo2021novel,tang2022gru}.
% Neural augmentation of Kalman Filters can also be employed to process measurements from multiple sensors such as GNSS and Inertial Measuring Units (IMU), before integrating them into a Kalman Filter \citep{guo2021novel,tang2022gru}.
% \hg{From Nir's paper, look at the end of page 8 (part IV-A external architectures). Citations 13, 27, 37. These preprocessing approaches are exactly what I am doing. https://arxiv.org/pdf/2410.12289}

\begin{figure*}[tbh!]
    \centering
    \begin{minipage}{0.25\textwidth}
        \centering
        \includegraphics[width=\columnwidth]{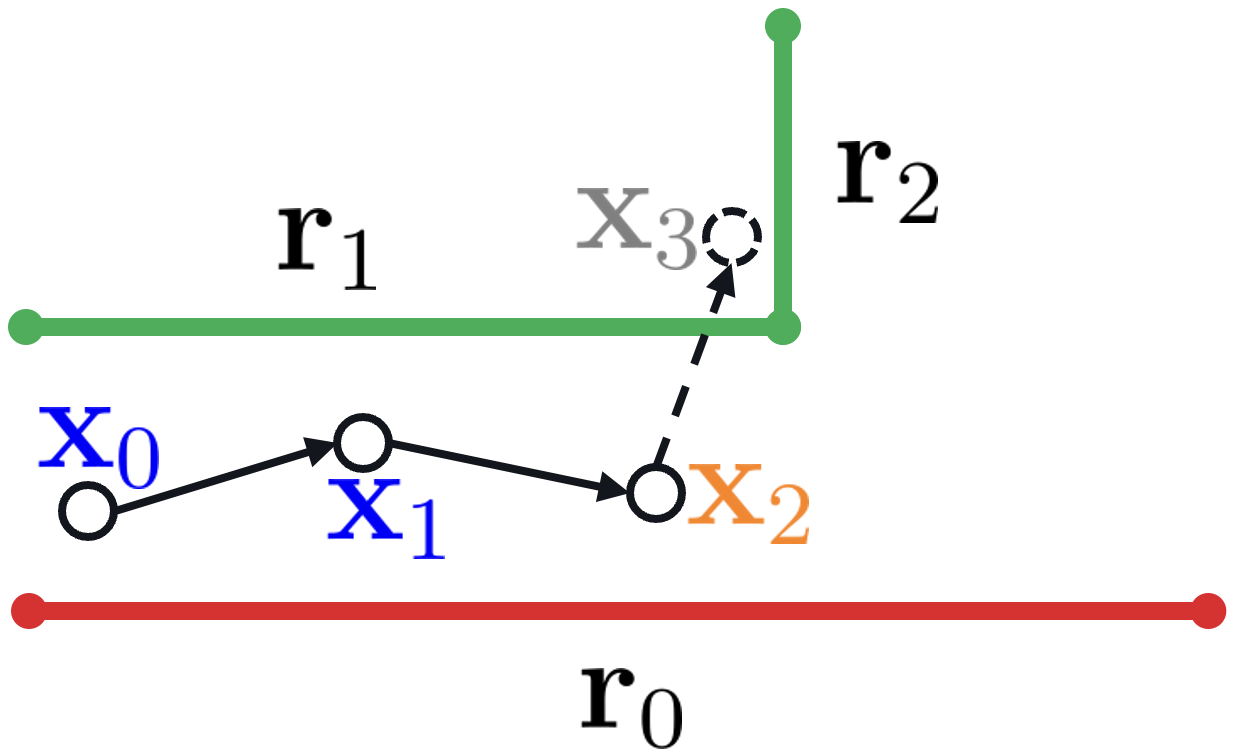}
        {\small ~\\ (a) Example scenario}\\
    \end{minipage}
    \begin{minipage}{0.03\textwidth}
    ~
    \end{minipage}
     \begin{minipage}{0.21\textwidth}
        \centering
        \includegraphics[width=\columnwidth]{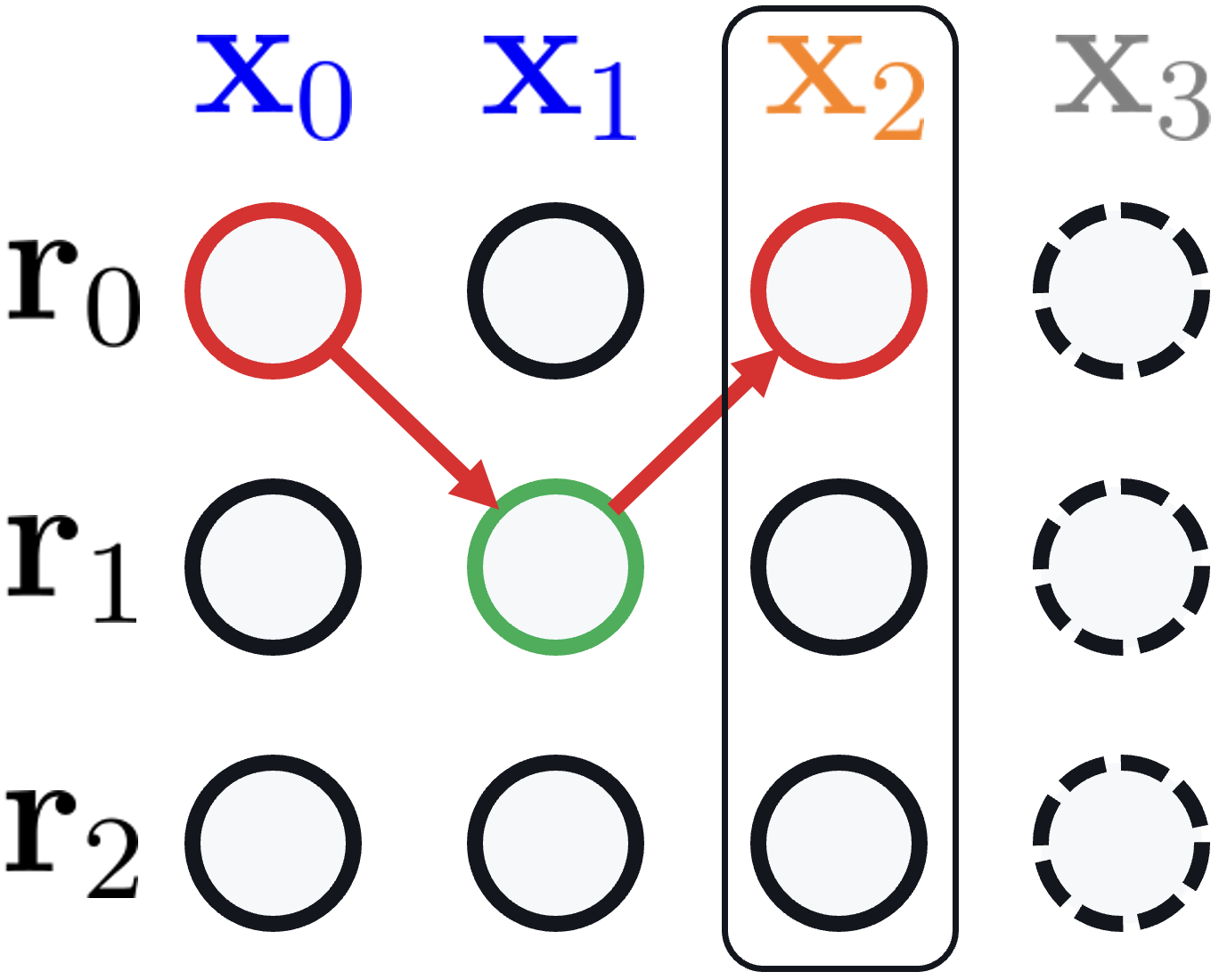}
        {\small (b) Instant}\\
    \end{minipage}
    \begin{minipage}{0.03\textwidth}
    ~
    \end{minipage}
     \begin{minipage}{0.21\textwidth}
        \centering
        \includegraphics[width=\columnwidth]{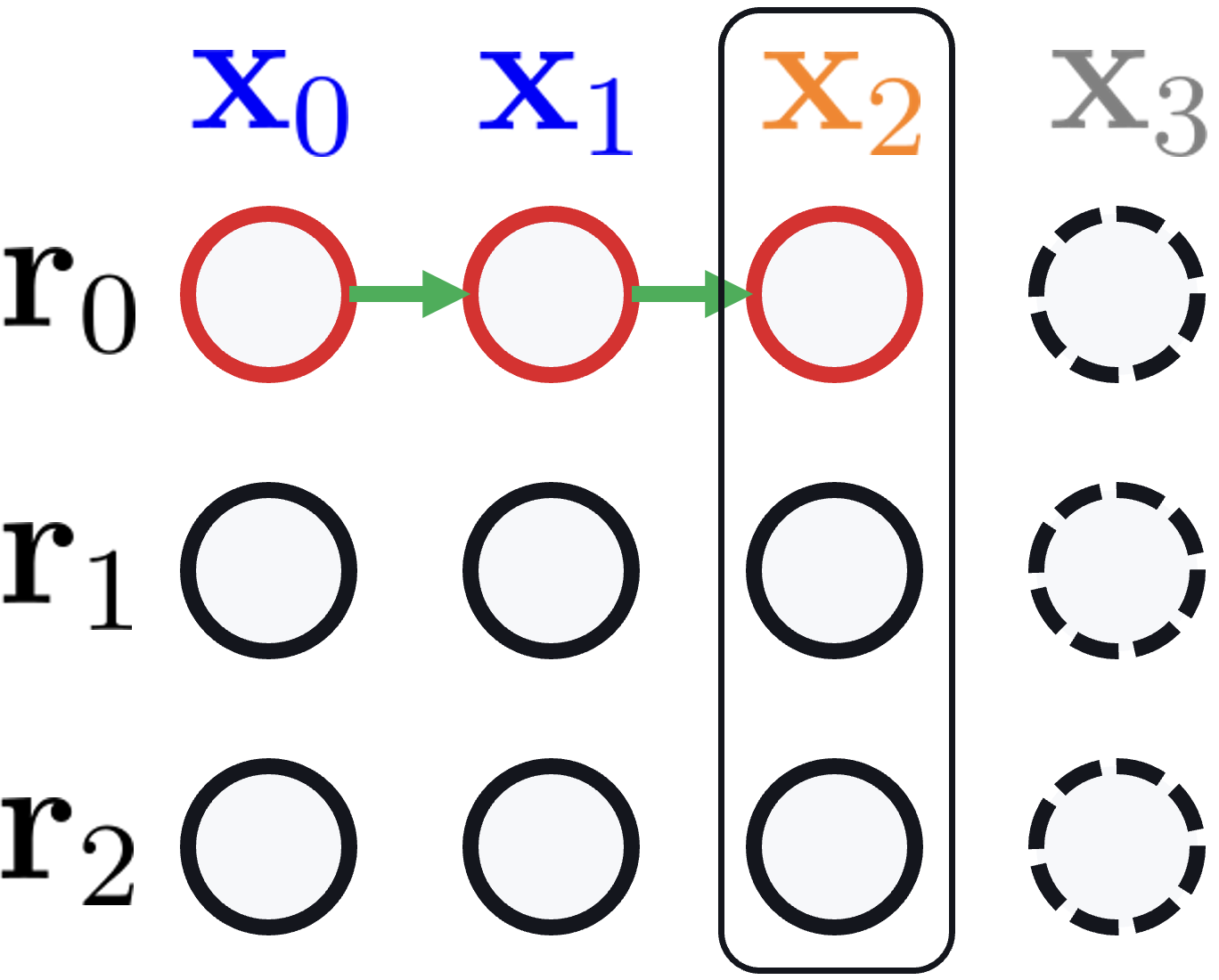}
        {\small (c) Viterbi}\\
    \end{minipage}
    \begin{minipage}{0.03\textwidth}
    ~
    \end{minipage}
     \begin{minipage}{0.21\textwidth}
         \centering
        \includegraphics[width=\columnwidth]{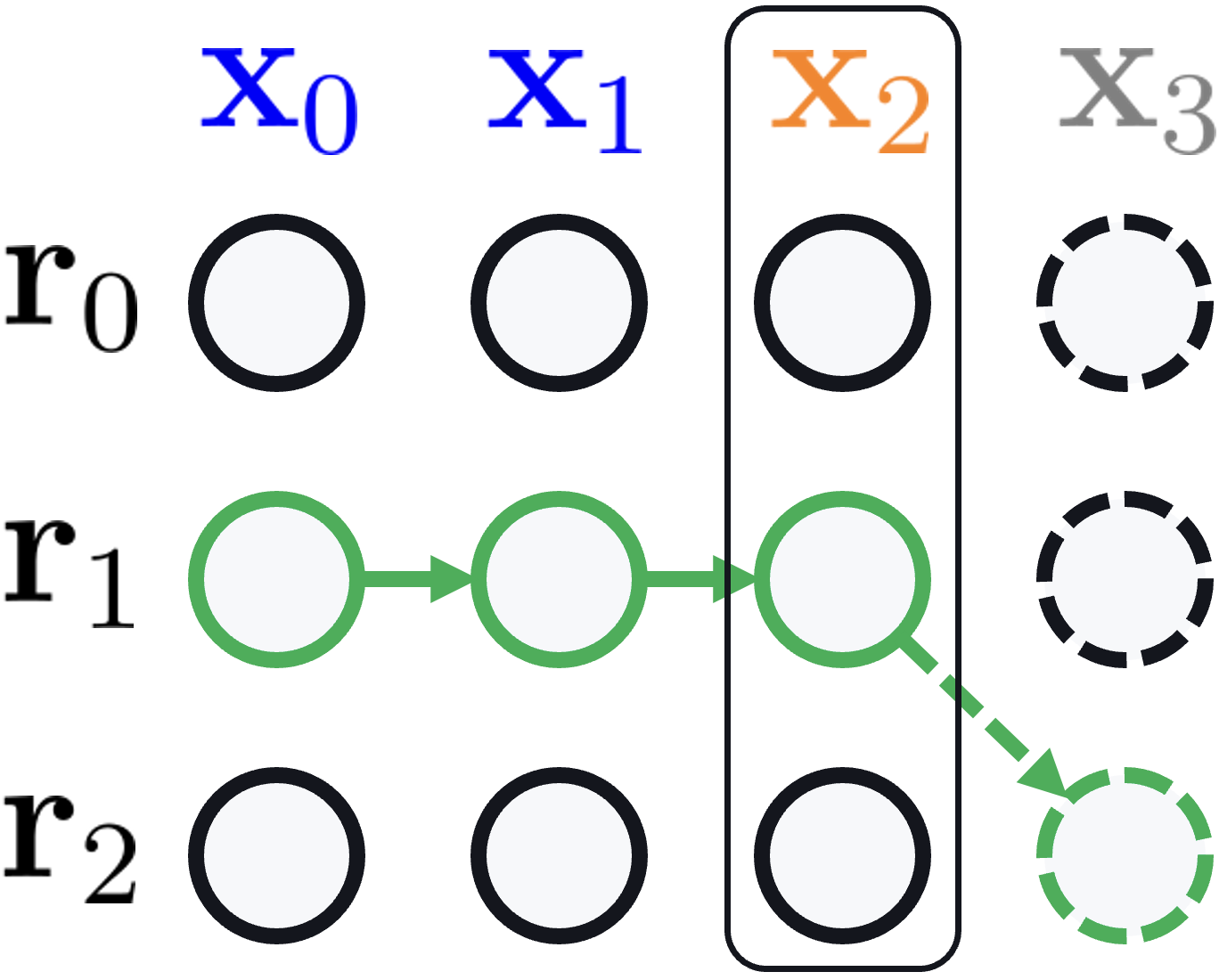}
        {\small (d) Bidirectional Viterbi}\\
    \end{minipage}
    \caption{Overview of different road selection algorithms. In (a) we show an example user trajectory estimated from noisy GNSS measurements, with past ($\color{blue} \mathbf{x}_0$, $\color{blue} \mathbf{x}_1$), current ($\color{orange} \mathbf{x}_2$), and the future ($\color{gray} \mathbf{x}_3$) locations. The road network has 3 segments including two parallel roads $\mathbf{r}_0$ and $\mathbf{r}_1$, with the true user trajectory following $\mathbf{r}_1$ and $\mathbf{r}_2$, marked in green. We compare three road selection methods, highlighting in green correct predictions for roads (nodes) and allowed transitions over the network (edges), while wrong ones are in red. An instantaneous solution (b) does not consider temporal relations, which might result in inconsistent predictions over time (jumping between similarly likely roads). Viterbi finds consistent solutions in terms of road connectivity, but can select the wrong path in case of noisy sequences. Bidirectional Viterbi (d) has access to future estimates and solves ambiguous cases by interpolating past and future trajectories.} 
    \label{fig:viterbi}
    
    \vspace{-0.5em}
\end{figure*}

\paragraph{GNSS positioning with Road Networks}
The impact of multipath effects on GNSS signals can be mitigated by incorporating additional sources of information, such as road networks. Road networks can be integrated in GNSS positioning systems without the need for additional sensors. This data can be limited to a description of the road or lane topology, or include additional features like road segment categories, driving speed, and directionality. 
\citet{quddus2015shortest} propose a genetic algorithm to select the most likely location among a set of candidate options based on local road network features. 
Other work jointly employ road network and GNSS data for the task of map matching, that is determining which sequence of road network segments best matches a GNSS measurement trajectory \citep{velaga2012map,feng2020deepmm,hu2023amm}. Map matching techniques only input information from the GNSS system to the map matching algorithm in order to select the most likely road. However, a feedback loop propagating road network information back into the GNSS positioning system could further improve its state and its ability to accurately process consecutive GNSS measurements. % No feedback loop to the GNSS system is used, which could in turn increase its accuracy in conse the next timestep by using the road network as an observation. %In this work, we specifically consider systems where such a feedback loop is used, i.e. the selected road is used as an observation for the GNSS tracking solution, see Fig. \ref{fig:figure_1}.
Thus, while map matching can be used to aid automated driving systems or model user driving behaviors, it does not provide a solution to the more general challenge of determining the exact user position on Earth at a given time. %\aj{previous sentence: I do not see why MM doesn't address determining user position.} 
%
%\db{by commenting out one paragraph, the following becomes disconnected from the previous one. I also don't like that we start talking about what is our method, and then we go back at discussing related work using road network.}
Other methods integrate road network information with KFs to improve the tracking performance over time. In this context, candidate locations on road segments are usually introduced to the KF as uni-modal Gaussian observations, to enable a direct update of the GNSS-based KF posterior. Features such as distance, heading, and speed alignment are used in heuristics or HMMs to determine the mean value of the distribution, while the standard deviation is typically estimated empirically and remains fixed during tracking \citep{el2005road,goh2012online,jagadeesh2017online,li2024novel}. In this paper, we demonstrate that the variance can be dynamically estimated for each timestep, thereby enabling more accurate localization when compared to a fixed variance. To the best of our knowledge, our method is also the first to propose a neural network to input road network data in a GNSS-based positioning system.
% Alternative approaches avoid the challenge of incorporating road segments as Gaussian observations by substituting Kalman Filters with Particle Filters (PF) \citep{davidson2010application,li2017using,kempinska2016probabilistic}. However, this approach can require tracking a large number of particles, which places a considerable computational burden on the system. % Furthermore, Particle Filters may encounter issues such as a low number of effective particles—where many particles are used to track the same state—or degeneracy, where only a single particle is effectively utilized.

% \aj{Here are some other relevant papers that may fit in the final story later:
% In channel tracking for wireless communication [Neural Augmentation of Kalman Filter with Hypernetwork for Channel Tracking] uses RNN to predict correction to all KF matrices
% In aviation [NNAKF: A Neural Network Adapted Kalman Filter for Target Tracking] uses RNN to estimate process noise covariance matrix (Q)
% In climate temperature prediction [Improving Accuracy of the Kalman Filter Algorithm in Dynamic Conditions Using ANN-Based Learning Module] uses MLP to predict measurement noise covariance (R)
% [A map matching algorithm based on modified hidden Markov model considering time series dependency over larger time span] removes the Markov assumption in HMM to have better tracking estimate. Their time window includes the future, therefore, their viterbi is non-causal and should produce a lag (not online)}

\section{Learning Road Network Selection}
% \mn{I feel that currently, the method section expects the reader to plow through a lot of formalism to understand what we did on a high-level. It is also confusing that it lays the foundation very precisely, but not only our contribution, some if it is just baseline, e.g. Viterbi road selection. So I either suggest to write a clear intro to section 3 that says "In 3.1 we introduce ..., In 3.2 we talk about our main contribution, etc" , and/or divide 3 into "Formalism" that defines the main equations, and highlights where the road selectors plug into, "Baselines" that describe instantaneous and causal Viterbi, but much shorter, with the equations being in appendix, and then "Our method" which does go into details. (Sub-section names were examples). I think 3.3 is good that it's separated from the rest of the NN, but name could be something like "Road Uncertainty prediction", or "Road Uncertainty prediction extension" or something along these lines. It confused me a bit that 3.2 title did not say "learning", 3.3 did, so I had different expectations from the sections when reading.} \hg{I have incorporated point 1 now. The and/or for 2 is still TODO. Personally, I am a bit hesitant to put all implementation details into the appendix, as it then becomes a bit of a `search' for how we did everything. But I also see the position that it is now a big info dump which also takes up a lot of space in the short 7-page format.} \db{I like the intro of Section 3 to explain the division of the subsections. I renamed Section 3.3 to include 'learning', for consistency with 3.4.}

In this section, we introduce a neural augmentation method for GNSS-based KF positioning to integrate the Road Network data in the KF measurement update. In Section \ref{sec:3.1}, we first formalize how to update the KF estimate given the road segment on which the user is located. In section \ref{sec:3.2} we explain how the correct road segment can be selected using the Viterbi algorithm. In section \ref{sec:3.3} we introduce our main contribution: a Temporal Graph Neural Network trained to select the correct road segment. Lastly, in Section \ref{sec:3.4}, we present an extension of our neural network to directly predict the road network covariance for the KF update.

\subsection{Integrating Road Data into the Kalman Filter}
\label{sec:3.1}
To ensure our solution can be applied to road data from different providers, we make minimal assumptions on the available information. We only describe the roads by their center-lane segment, without relying on lane-level information, which is often incomplete, outdated or unavailable. For the same reasons, we treat road features such as driving speed and directionality as optionally available. We describe the dataset in more detail in Section \ref{sec:exp_setup}.

For the purpose of road segment selection it is convenient to consider a graph representation in which the nodes are the road segments, and the edges are intersection or curvature points connecting two or more road segments in the network. We therefore define a road network graph as $\mathcal{G} = (\rmR, \rmA)$ with $\rmR \in \mathbb{R}^{\text{N} \times \text{D}}$ describing $\text{D}$-dimensional features for each of the $\text{N}$ road segments, and $\rmA \in \mathbb{R}^{\text{N} \times \text{N}}$ being a binary adjacency matrix representing the network structure.

Road network information can be used to post-process a KF estimate and generate an updated positioning prediction, for example by snapping the KF position on a selected road segment. Alternatively, the selected road segment could be directly integrated in the KF through an additional measurement update, as show in Figure \ref{fig:figure_1}. In this work, we consider a KF with mean $\rvx \in \mathbb{R}^{8}$ and covariance $\rmP \in \mathbb{R}^{8 \times 8}$ to track the user position $\rvx_{\text{pos}} \in \mathbb{R}^{3}$ and velocity $\rvx_{\text{vel}} \in \mathbb{R}^{3}$ on the local geodetic coordinate system as well as the receiver clock bias and drift $\rvx_{\text{clock}} \in \mathbb{R}^{2}$.
The road network can be employed as a prior for the positioning task by updating the KF state based on the best matching road segment $r^*$. The road segment can be selected among a set of candidate segments as the one minimizing an arbitrary cost function $J(r_i)$ as: 
\begin{equation}
\label{eq:arg}
r^* = \argmin_{r_i} J(r_i)  \quad\operatorname{with} \quad0\leq i<N ,
\end{equation}

A simple baseline for segment selection (see Figure \ref{fig:viterbi}b) can be devised by defining the cost function in Equation \ref{eq:arg} as the distance between the estimated user position $\rvx_{\text{pos}}$ and a road segment $r_i$:
\begin{equation}
    J_{\text{pos}}(r_i) = \operatorname{dist}(\rvx_{\text{pos}},r_i),
    \label{eq:emission_position}
\end{equation}

where $\operatorname{dist}(.,~.)$ is the Euclidean distance between a point and a line segment in meters. We refer to this baseline as \emph{Instant} in our experiments, as the selection only considers the current position estimate, and not the previous trajectory of the user.
Given the KF posterior (denoted with $^+$) after the GNSS update step with mean $\rvx_{\operatorname{GNSS}}^+$ and covariance $\rmP_{\operatorname{GNSS}}^+$, the road network measurement update is:

\begin{align}
\rmK &= \rmP_{\operatorname{GNSS}}^+ \rmH^{\operatorname{T}} \left( \rmH \rmP_{\operatorname{GNSS}}^+ \rmH^{\operatorname{T}} + \rmV \right)^{-1}, \nonumber \\
\rvx_{\operatorname{RN}}^+ &= \rvx_{\operatorname{GNSS}}^+ + \rmK \left(\rvz - \rmH \rvx_{\operatorname{GNSS}}^+ \right), \label{eq:kf_update} \\
\rmP_{\operatorname{RN}}^+ &= \rmP_{\operatorname{GNSS}}^+ - \rmK \rmH \rmP_{\operatorname{GNSS}}^+, \nonumber
\end{align}

% \aj{In the literature, Q is used for process noise covariance matrix in KF. I assume the usual R was not used here because the road features are shown by R. I would change Q to R and R to F for example, but I can imagine that being a pain as it is everywhere in the paper. I would at least change this Q to some other letter.} \hg{I agree, I have now changed it to $\rm\Sigma$ for covariance matrix. Waht do you think of this?}\db{sigma is also used occasionally instead of P, as the state covariance. It is also a bit odd to have sigma for covariance but not mu for mean. I prefer some other variable name. If you like to avoid R, we could use U, V, B or some other unused letter.}\hg{Changed to $\rmV$ for variance.} 
where $\rvz$, $\rmV$ and $\rmH$ are respectively the mean, covariance and observation matrix for the road network measurement, $\rmK$ is the Kalman gain, $\rvx_{\operatorname{RN}}^+$ and $\rmP_{\operatorname{RN}}^+$ are the KF posterior mean and covariance after the road network update step, and $^{\operatorname{T}}$ denotes the transpose operator. Appendix \ref{app:equations}
includes the complete derivations showing how $\rvz$, $\rmV$ and $\rmH$ are derived from the road network position and heading.

\subsection{Road Selection with the Viterbi Algorithm}
\label{sec:3.2}
The instantaneous selection of road segments often results in unstable and inaccurate predictions, as it does not capture temporal patterns over consecutive time-steps. One solution posed in literature is the use of a Hidden Markov Model to process the vehicle trajectory over the graph. In such an HMM, the observation is the history of estimated user positions, and the latent variable is the road segment on which the user is currently located. The Viterbi algorithm can be used online to determine the maximum a posteriori as the most likely sequence of road segments (Figure \ref{fig:viterbi}c).

To create a non-learnable baseline for road selection with Viterbi, we consider a HMM in which each road segment is a state, and the transition and emission probability functions are defined considering the relation between the road network and the estimated user location from the KF. 
%For the emission probability calculation of each state, we calculate a cost for the user position and heading first, which we then transform into an emission probability density function (PDF). 
The emission probability for each road segment is computed based on a position and heading cost.
The position cost $J_{\text{pos}}$ is defined as in Equation \ref{eq:emission_position}, and the heading cost is calculated as: 
% $J_{\theta}(r_i) = 1 - \left|\cos(\theta_{\rvx} - \theta_{r_i})\right|$,
\begin{equation}
    J_{\theta}(r_i) = 1 - \left|\cos(\theta_{\rvx} - \theta_{r_i})\right|,
    \label{eq:emission_heading}
\end{equation}
where $\theta_{\rvx}$ is the user heading and $\theta_{r_i}$ is the road heading. With the absolute value we allow matching both directions along the road. The information about one-way road directionality, if available, will be encoded in the transition probabilities. Finally, the emission PDF is calculated using a weighted average of the heading and positions costs, as:
\begin{equation}
    p(\rvx|r_i) = \max \left(1 - \cfrac{ \beta J_{\text{pos}}(r_i) + J_{\theta}(r_i)}{2},~\epsilon \right),
    \label{eq:emission_total}
\end{equation}
where $\beta$ is a hyper-parameter to balance between the two cost functions, and $\epsilon=0.01$ ensures that all road segments retain a non-zero emission probability.
The transition probabilities between the different road segments are simply modeled with the PDF:
\begin{equation}
    p(r_i|r_j) = \begin{cases}
            1 & \text{if~~} r_i \in \text{~k-Hop}(r_j; k) \\
            0 & \text{otherwise}
    \end{cases},
    \label{eq:transition}
\end{equation}
where the function $\text{k-Hop}(r_j; k)$ defines the $k$-hop graph neighborhood of $r_j$, which is a set including each road $r_i$ that the user can reach within a maximum of $k$ intersections starting from $r_j$, and taking one-way directionality into account. Setting $k > 1$ is useful to handle cases in which the vehicle is moving at high speed, traversing multiple road segments in a single time-step.

\begin{figure}[t!]
    \centering
    \includegraphics[width=0.96\linewidth]
    {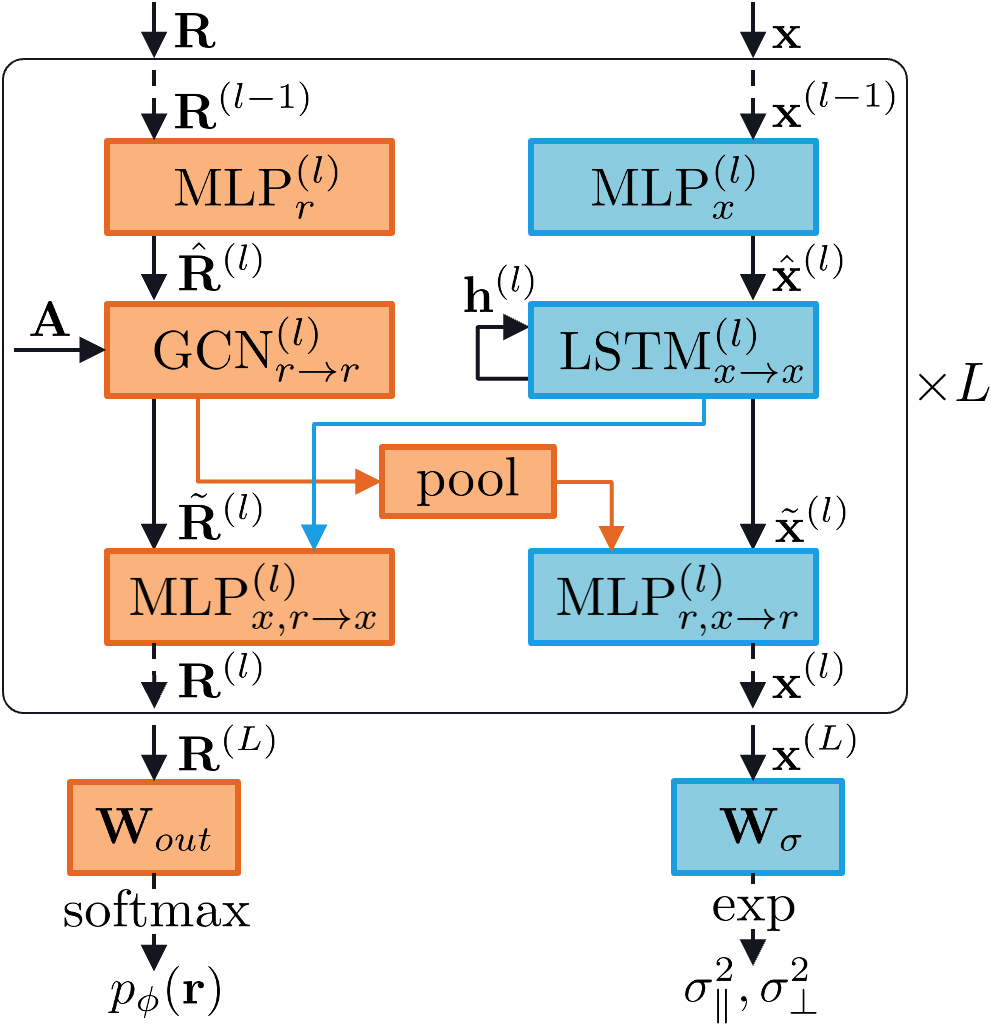}    
    \vspace{-0.5em}
    \caption{Architecture of the proposed TGNN.}
    \label{fig:tgnn}
    
    \vspace{-0.8em}
\end{figure}

The Viterbi algorithm can efficiently run online % to determine the maximum probability to reach a given road segment at the current time-step $t$
through the recursive PDF formulation:
\begin{equation}
\resizebox{.88\linewidth}{!}{$\displaystyle
    p(r_i,t) = \begin{cases}
            p(\rvx_{\text{pos}}|r_i) & \text{if~~} t = 0 \\
            \max\limits_{r_j} ~ p(r_j,t\!-\!1) ~ p(r_i|r_j) ~ p(\rvx_{\text{pos}}|r_i) & \text{if~~} t > 0 \\ % simplified from \max\limits_{r_j, 0\leq j<N} p(r_i,t-1)
    \end{cases}
$},
\end{equation}

where at timestep $t = 0$ we assume a uniform prior over all roads. Therefore, the most likely road segment at the current time-step $t$ can be selected according to Equation \ref{eq:arg} with the Viterbi cost function $ J_{\text{Viterbi}}(r_i) =- p(r_i,t)$.

% \begin{equation}
% J_{\text{Viterbi}}(r_i) =- p(r_i,t)
% \label{eq:select_viterbi}.
% \end{equation} 
We refer to this method as \emph{Viterbi} in our experiments.

%-------------------------------------------------------------------------%
\subsection{Road Selection with a Trained TGNN}
\label{sec:3.3}
If future position estimates were available, a \enquote{bidirectional} Viterbi algorithm could be used to find a smoother trajectory on the road network, temporally consistent with both past and future observations (see Figure \ref{fig:viterbi}d). We refer to this bidirectional Viterbi algorithm as the \emph{Oracle} method in our experiments, since it relies on future observations which are not available during online inference.

% \todo{choose a name for our architecture? Temporal Road Graph Network? TRGN?}\db{I think we converged into TGNN? Do you want to discuss alternatives or happy with this? Main choice is between coming up with a custom name for our specific solution/architecture, versus a more general naming as the current one. Both have pros and cons.} \hg{I quite like the TGNN naming convention. It also makes the ablation simpler as GNN and MLP (could even be NN). But it does mean that there is no custom name for it. Lets discuss online.}

While future information is not available in practice, we propose training a neural network to learn to predict the optimal solutions found by the \emph{Oracle}. We implement this as a novel Temporal Graph Neural Network architecture for road segment prediction. We choose Graph Neural Networks (GNNs) \citep{scarselli2008graph,kipf2016semi,zhou2020graph} as they can process structured graphs with varying number of nodes and are invariant to the nodes order. Temporal Graph Neural Networks \citep{zhao2019t,cao2020spectral,rossi2020temporal} are a specific type of GNNs which can also process temporal relations in the graph. We show in Figure \ref{fig:tgnn} the architecture of the proposed TGNN.

Our model $\mathbf{\phi}$ takes as input the KF state $\mathbf{x}$, the road segment features $\mathbf{R}$, and the road network adjacency matrix $\mathbf{A}$, and outputs a probability for each road segment:
\begin{equation}
P_{\mathbf{\phi}}(\mathbf{r}) = \mathbf{\phi}\left(\mathbf{x}, \mathbf{R}, \mathbf{A} \right)
\end{equation}
% \begin{equation}
% r^* = \argmin_{r_i} \operatorname{dist}(\rvx_{\text{pos}},r_i)
% \end{equation}
% \begin{equation}
% \resizebox{.88\linewidth}{!}{$\displaystyle
% r^* = \argmax_{r_i} \begin{cases}
%             p(\rvx_{\text{pos}}|r_i) & \text{if~~} t = 0 \\
%             \max\limits_{r_j} ~ p(r_j,t\!-\!1) ~ p(r_i|r_j) ~ p(\rvx_{\text{pos}}|r_i) & \text{if~~} t > 0 \\ % simplified from \max\limits_{r_j, 0\leq j<N} p(r_i,t-1)
%     \end{cases}
% $}
% \end{equation}
% \begin{equation}
% r^* = \argmax_{r_i} P_{\mathbf{\phi}}\left(r_i | \mathbf{x}, \mathbf{R}, \mathbf{A} \right)
% \end{equation}

The neural network consists of a sequence of $L$ repeated blocks which include feature transformation, local message passing and cross message passing layers. The user-level features $\mathbf{x}$ and road-level features $\mathbf{R}$ are processed on two distinct paths throughout the network, only merging in the cross message passing layers. 
In each block $l$, the feature transformation layers are implemented as two simple MLPs projecting the input features:
% \begin{align}
% \hat{\rvx}^{(l)} = \operatorname{MLP}^{(l)}_x\left(\rvx^{(l-1)}\right), \quad \hat{\rmR}^{(l)} = \operatorname{MLP}^{(l)}_r\left(\rmR^{(l-1)}\right) .
% \end{align}
\begin{equation}
\resizebox{.88\linewidth}{!}{$\displaystyle
    \hat{\rvx}^{(l)} = \operatorname{MLP}^{(l)}_x\left(\rvx^{(l-1)}\right), \quad \hat{\rmR}^{(l)} = \operatorname{MLP}^{(l)}_r\left(\rmR^{(l-1)}\right) .
$}
\end{equation}
Local message passing layers consist of a Graph Convolutional Network (GCN) \citep{kipf2016semi} layer to propagate information across the road graph ($r\rightarrow r$), and a Long Short-Term Memory (LSTM) \citep{hochreiter1997long} layer to capture temporal patterns ($x\rightarrow x$) in the history of KF states:
\begin{align}
\tilde{\rvx}^{(l)}, \rvh^{(l)}_t &= \operatorname{LSTM}^{(l)}_{x\rightarrow x}\left(\hat{\rvx}^{(l)}, \rvh^{(l)}_{t-1}\right), \\
% \tilde{\rvx}^{(l)} &= \operatorname{LSTM}^{(l)}_{x\rightarrow x}\left(\hat{\rvx}^{(l)}\right), \\
\tilde{\rmR}^{(l)} &= \operatorname{GCN}^{(l)}_{r\rightarrow r}\left(\hat{\rmR}^{(l)}, \rmA \right).
\end{align}
% \begin{equation}
% \resizebox{.88\linewidth}{!}{$\displaystyle
%     \tilde{\rvx}^{(l)} = \operatorname{LSTM}^{(l)}_{x\rightarrow x}\left(\hat{\rvx}^{(l)}, \rvh^{(l)}_{t-1}\right), 
%     \quad    \tilde{\rmR}^{(l)} = \operatorname{GCN}^{(l)}_{r\rightarrow r}\left(\hat{\rmR}^{(l)}, \rmA \right).
% $}
% \end{equation}
%
Finally, the cross message passing layers consist of MLPs which update the user-level features given the road-level features ($x, r\rightarrow x$), and vice-versa ($r, x\rightarrow r$):
\begin{align}
\rvx^{(l)} &= \operatorname{MLP}^{(l)}_{x, r\rightarrow x}\left(\left[ \tilde{\rvx}^{(l)}, \operatorname{mean\_pool}\big(\tilde{\rmR}^{(l)}\big)\right] \right), \\
\rmR^{(l)} &= \operatorname{MLP}^{(l)}_{r, x\rightarrow r}\left(\left[ \tilde{\rmR}^{(l)}, \tilde{\rvx}^{(l)} \right]\right).
\end{align}
Mean pooling is used to summarize the road-level feature vectors in a single feature vector of fixed dimensions.
After $L$ blocks, a linear layer $\rmW_{out}$ projects the road-level features into logits and a softmax function is used to convert these to probabilities: $P_{\mathbf{\phi}}(\rvr) = \operatorname{softmax}\left( \rmW_{out} \rmR^{(L)} \right)$.

We train the model using a cross-entropy loss to match the predicted probabilities $p_{\mathbf{\phi}}(\mathbf{r})$ with the road segments $r^{*}_{\operatorname{oracle}}$ selected by the bidirectional Viterbi \emph{Oracle}: 
\begin{align}
L_{\text{CE}} = \operatorname{CE}(r^{*}_{\operatorname{oracle}}, P_{\mathbf{\phi}}(\rvr)).
\end{align}

Finally, the most likely road segment is selected as in Equation \ref{eq:arg} based on the cost function $J_{\text{TGNN}}(r_i) =- P_{\mathbf{\phi}}(r_i)$.
% \begin{equation}
% \label{eq:select_tgnn}
% J_{\text{TGNN}}(r_i) =- P_{\mathbf{\phi}}(r_i).
% \end{equation} 
% \aj{maybe we can give a name to the methods explained in 3.1 and 3.2 at the end of these sections and use the same name in results section?} \db{Added. Hans please confirm is this is good with you or not.} \hg{Yeah, looks good!}

% \begin{figure}
%     \centering
%     \includegraphics[width=1\linewidth]{old_figures/viterbi_example.pdf}
%     \caption{Example of the road segments selected by the Viterbi algorithm that runs on the entire trajectory of ground truth user locations.}
%     \label{fig:viterbi_example}
% \end{figure}

% \db{do we need this paragraph and plot? don't think so.}
% Figure \ref{fig:viterbi_example} shows an example of what the selected road segments for a given drive look like. As can be seen from the figure, the Viterbi algorithm sometimes extracts the wrong segment. Future research could look into improving this selection. For now, we simply calculate the distance between the original ground truth locations and the selected Viterbi segments. If it is larger than 20 meters, we do not use the segment for training or to calculate selection accuracy. 

%-------------------------------------------------------------------------%
\subsection{Learning to Predict the Road Uncertainty}
\label{sec:3.4}
The road network KF update described in Equation \ref{eq:kf_update} requires not only the mean $\rvz$ of the selected road segment, but also its covariance $\rmV$.
The covariance is a diagonal matrix with two non-zero values: the variance parallel to the road ($\sigma_{\parallel}^2$) and the variance perpendicular to the road ($\sigma_{\perp}^2$). 

While these parameters can be tuned offline using a calibration set, we instead equip our neural network with an additional output head to predict them online. This allows the TGNN to dynamically adapt its confidence in the road network prediction depending on the current scenario, for example predicting higher uncertainty in case of cluttered or ambiguous road network structure, and lower uncertainty otherwise. The uncertainty prediction head is implemented as a linear projection head on top of the features at the last layer of TGNN:
\begin{equation}
    \left[\sigma_{\parallel}^2, \sigma_{\perp}^2\right] = \operatorname{exp}\left(\rmW_{\sigma} \rvx^{(L)}\right).
\end{equation} 

To train the network for uncertainty prediction we use the (MSE) loss between the output of the KF and the ground truth user state. This is possible because the KF is fully differentiable \citep{shlezinger2024ai}. The complete loss function used to train the TGNN is a combination of the road uncertainty and road selection losses, with a parameter $\lambda$ to balance the two components:
\begin{align}
    L_{\text{MSE}} &= \operatorname{MSE}(\rvx_{\text{gt}}, \rvx_{\operatorname{RN}}^+). \\
    L &= L_{\text{CE}} + \lambda L_{\text{MSE}},
\end{align}

% \db{separated the definition of the loss components (CE is now defined earlier) and corrected the variable names to be consistent throughout the paper.}

% \todo{Hans to write this paragraph and equation (see notation in previous ones).}
% standard deviation needs to be hardcoded. we can learn it too in a supervised setup. describe the two std components. 
% \begin{equation}
% equation.
% \end{equation}
% mention the loss function used to train this through the KF equations.
% Briefly mention we could also learn directly the measurement, but that is much harder than selection, which we found performing better.

% The TGNN is then used during online inference to predict the road network measurement and its uncertainty matrix which is then used for the Kalman Filter update:

\section{Results}

\subsection{Experimental Setup}
\label{sec:exp_setup}
% \hg{This section needs careful checking to ensure it does not breach CCI} \db{Amir can you check this?} \aj{Checked this. The info shared seems to be already included in previous paper. I only added one comment}

% \begin{table*}
% \centering
% \begin{tabular}{lllcc}
% \toprule
% Tracking  & Road Selection & Road Covariance & HE@50$^{th}$                             & HE@95$^{th}$                             \\
% \midrule
% LS       & -              & -               & 20.43\phantom{$~\pm~$0.00}              & 115.97\phantom{$~\pm~$0.00}             \\
% KF       & -              & -               & 10.75\phantom{$~\pm~$0.00}              & \phantom{0}77.23\phantom{$~\pm~$0.00}   \\
% \midrule
% KF       & Genie          & 0               & \phantom{0}3.72\phantom{$~\pm~$0.00}    & \phantom{0}11.40\phantom{$~\pm~$0.00}    \\
% \midrule
% KF       & Instant        & Grid Search     & \phantom{0}7.96\phantom{$~\pm~$0.00}    & \phantom{0}68.86\phantom{$~\pm~$0.00}   \\
% KF       & Viterbi        & Grid Search     & \phantom{0}8.02\phantom{$~\pm~$0.00}    & \phantom{0}68.27\phantom{$~\pm~$0.00}   \\
% \textbf{KF}       & \textbf{TGNN} & \textbf{TGNN} & \textbf{\phantom{0}8.74$~\pm~$0.15}              & \textbf{\phantom{0}55.02$~\pm~$2.21}  \\
% \bottomrule
% \end{tabular}
% \caption{\label{tab_main_results} Quantitative results comparing our methods to the baselines. The top two rows do not make use of the road measurement at all, while the third row makes use of the genie information.} 
% \end{table*}

\begin{table}
\caption{\label{tab_main_results} Quantitative results comparing our method (in \textbf{bold}) to different baselines. The first two baselines do not use road network data, while the \emph{Oracle} makes use of future information. 
% \db{Alternative table for Table 1 which spans a single column. we might want to shrink to half columns if possible, since it is quite narrow for full page.} \hg{Agreed, this looks much better, also considering the other tables are single column. I have updated the other tables to look like this one.}
} 
\centering
\setlength{\tabcolsep}{5pt}
\begin{tabular}{lllll}
\toprule
 & \begin{tabular}[c]{@{}l@{}}Road \\ Select.\end{tabular}  & \begin{tabular}[c]{@{}l@{}}Road \\ Covariance\end{tabular} & 
 \begin{tabular}[l]{@{}l@{}}HE@50$^{th}$ \\ \hspace{0cm}[m]\end{tabular}                             & \begin{tabular}[l]{@{}l@{}}HE@95$^{th}$ \\ \hspace{0cm}[m]\end{tabular}  \\
 % HE@50$^{th}$ [m]                             & HE@95$^{th}$ [m]                  \\
\midrule
LS       & -              & -               & 20.43\phantom{\scriptsize{$~\pm~$0.00}}            & 115.97\phantom{\scriptsize{$~\pm~$0.00}}             \\
KF       & -              & -               & 10.75\phantom{\scriptsize{$~\pm~$0.00}}             & \phantom{0}77.23\phantom{\scriptsize{$~\pm~$0.00}}    \\
\midrule
KF       & Oracle          & 0               & \phantom{0}3.72\phantom{\scriptsize{$~\pm~$0.00}}     & \phantom{0}11.40\phantom{\scriptsize{$~\pm~$0.00}}     \\
\midrule
KF       & Instant        & Grid Search     & \phantom{0}7.96\phantom{\scriptsize{$~\pm~$0.00}}     & \phantom{0}68.86\phantom{\scriptsize{$~\pm~$0.00}}   \\
KF       & Viterbi        & Grid Search    & \phantom{0}8.02\phantom{\scriptsize{$~\pm~$0.00}}    & \phantom{0}68.27\phantom{\scriptsize{$~\pm~$0.00}}    \\
\textbf{KF}       & \textbf{TGNN} & \textbf{TGNN} & \phantom{0}8.74\scriptsize{$~\pm~$0.15}              & \phantom{0}55.02\scriptsize{$~\pm~$2.21}  \\
\bottomrule
\end{tabular}
\vspace{-0.2em}
\end{table}
\begin{table}[t]
\centering
\caption{Ablating road selection or covariance prediction with non-learnable components (Viterbi or grid search, respectively).}
\setlength{\tabcolsep}{5.5pt}
\begin{tabular}{lllll}
\toprule
 & \begin{tabular}[c]{@{}l@{}}Road \\ Select.\end{tabular}  & \begin{tabular}[c]{@{}l@{}}Road \\ Covariance\end{tabular} & 
 \begin{tabular}[l]{@{}l@{}}HE@50$^{th}$ \\ \hspace{0cm}[m]\end{tabular}                             & \begin{tabular}[l]{@{}l@{}}HE@95$^{th}$ \\ \hspace{0cm}[m]\end{tabular}  \\
 %HE@50$^{th}$                             & HE@95$^{th}$                  \\
 \midrule
KF       & Viterbi        & TGNN                  & 8.69\scriptsize{$~\pm~$0.18}            & 67.08\scriptsize{$~\pm~$3.49}   \\
KF       & TGNN           & Grid Search           & 8.36\scriptsize{$~\pm~$0.38}            & 63.72\scriptsize{$~\pm~$9.00}    \\
\textbf{KF}       & \textbf{TGNN} & \textbf{TGNN} & 8.74\scriptsize{$~\pm~$0.15}   & 55.02\scriptsize{$~\pm~$2.21}  \\
\bottomrule
\end{tabular}
\label{tab_selection_ablations}
\end{table}

\paragraph{Dataset}
In this manuscript we employ the real-world GNSS dataset introduced by \citet{ml2}. The data consists of multiple unconnected drives from four cities in different countries. For each drive, we have access to the ground-truth user location as well as GNSS pseudo-range measurements corrected for known error terms. We supplement the GNSS dataset with road network information using the open-source data provided by \citet{OpenStreetMap}. The road network topology is described with an undirected road-level graph, where nodes represent either intersections or a curvature point in the road. Each road segment describes the center of the road, while the positions and offsets of multiple lanes in the road are not included. The following features are additionally paired to each road segment: coordinates of the two end points, segment length, number of lanes, maximum driving speed, roadway type (highway, primary, residential, etc.), and whether it is a one-way street. This information is not always present, in which case we set it to a default value when using as a neural network input. Other input features include the KF mean and uncertainty, and the probabilities from the Viterbi algorithm. The dataset is split in three folds with no regional overlap, and a leave-one-out cross-validation method is used in our experiments, averaging the scores across the holdout folds. Each fold includes both open-sky and challenging urban scenarios.
We provide in Appendix B %\ref{app:dataset} 
additional details on the processing of GNSS and Road Network data, as well as the complete list of features used as input to the \emph{TGNN}.

\paragraph{Evaluation Metrics}
We evaluate the proposed methods and baselines using the cumulative distribution function (CDF) of the Horizontal Error (the distance between the predicted and target location). Specifically, we report results for the 50\textsuperscript{th} percentile (median) and the 95\textsuperscript{th} percentile of the CDF, with the latter being the primary metric. The horizontal error at higher percentiles reflects the performance in challenging scenarios where significant errors occur. These errors, in the order of $\sim60m$ , can severely impact downstream applications of GNSS positioning, such as lane estimation and side-of-the-street detection. Slight changes in HE@50, which is usually under $\sim10$ meters, are relatively less important in this context. 
For each result from a learnable method, we report mean KPI and standard deviation over 10 seeds.

\begin{table}[t]
\centering
\caption{\label{tab_network_ablations} Comparison between the \emph{TGNN} architecture against the ablated variants \emph{GNN} (without temporal LSTM) and  \emph{MLP} (without graph convolution and LSTM).} 

\setlength{\tabcolsep}{11.5pt}
\begin{tabular}{lll}
\toprule
Neural Network ~~~~ & 
 \begin{tabular}[c]{@{}l@{}}HE@50$^{th}$ \\ \hspace{0cm}[m]\end{tabular}                             & \begin{tabular}[c]{@{}l@{}}HE@95$^{th}$ \\ \hspace{0cm}[m]\end{tabular}  \\
 %HE@50$^{th}$                             & HE@95$^{th}$                  \\
\midrule
MLP             & 9.13\scriptsize{$~\pm~$0.31}              & 59.68\scriptsize{$~\pm~$3.94}             \\
GNN             & 8.82\scriptsize{$~\pm~$0.17}              & 56.90\scriptsize{$~\pm~$2.83}            \\
\textbf{TGNN}   & 8.74\scriptsize{$~\pm~$0.15}     & 55.02\scriptsize{$~\pm~$2.21}  \\
\bottomrule
\end{tabular}
\vspace{-0.2em}
\end{table}

\begin{table}[t!]
\centering
\caption{\label{tab_feature_ablation} Effects of using subsets of the available input features with the proposed TGNN method. } 
\setlength{\tabcolsep}{5.5pt}
\begin{tabular}{lll}
\toprule
Features       &                                         \begin{tabular}[c]{@{}l@{}}HE@50$^{th}$ \\ \hspace{0cm}[m]\end{tabular}                             & \begin{tabular}[c]{@{}l@{}}HE@95$^{th}$ \\ \hspace{0cm}[m]\end{tabular}  \\
 %HE@50$^{th}$                             & HE@95$^{th}$                  \\
\midrule
Viterbi Prior                                                                   & 10.16\scriptsize{$~\pm~$0.15}                          & 72.79\scriptsize{$~\pm~$1.21} \vspace{3pt} \\ 
Viterbi Prior + Distances                                                       & \phantom{0}8.64\scriptsize{$~\pm~$0.17}                & 61.43\scriptsize{$~\pm~$3.52} \vspace{3pt} \\ 
\begin{tabular}[c]{@{}l@{}}Viterbi Prior + Distances + \\ Road Type + Max Speed\end{tabular}    & \phantom{0}8.65\scriptsize{$~\pm~$0.21}                & 56.29\scriptsize{$~\pm~$1.66} \vspace{3pt} \\ 
\textbf{All}                                                        & \phantom{0}8.74\scriptsize{$~\pm~$0.15}       & 55.02\scriptsize{$~\pm~$2.21} \vspace{2pt} \\
\bottomrule
\end{tabular}
\vspace{-0.5em}
\end{table}

\paragraph{Methods and baselines}
% \todo{Hans to add main details about KF initialization, baselines and NN architectures, e.g.: values for all hyper-parameters in the NN and optimization. See example template below on the optimization part. Part of this might go to appendix if we need space}
The KF is initialized using the least squares method described in \citet{ml2}. We evaluate four different strategies to select the road segment for the KF update: \emph{Instant} and \emph{Viterbi} as baselines, the bidirectional Viterbi \emph{Oracle}, and our proposed \emph{TGNN} solution. 
Appendix \ref{app:hyper} reports
all implementation details and selected hyper-parameters to reproduce our experiments. Note that the existing methods discussed in Section \ref{sec:related} are either heuristics for KF snapping similar to our \emph{Instant} baseline, or consider the task of map matching instead of user positioning, and are therefore not suitable baselines for our experiments.

\subsection{Quantitative Results}
We report the end-to-end positioning error averaged over folds for different positioning algorithms in Table \ref{tab_main_results}, where LS refers to the Least Squares algorithm for instantaneous positioning. %\aj{simple techniques "can" use road network: I cannot imagine the simple method "instant" not using the road network. Maybe something like: using road networks on top of KF or HMM already improves ...?}\db{I think this meant to say that Road Networks can be used to improve KF even without need for NN training. Maybe we can remove the `model-based` part if confusing. Indeed neural networks are also often referred to as `models`.} \aj{My suggested sentence: We found that even simple model-based techniques that use the road network data (\emph{Instant} and \emph{Viterbi}) improve the prediction with a 10 meters decrease in localization error at the 95$^{th}$ percentile when compared against the GNSS-based KF.} 
We find that even simple techniques that use road network data (\emph{Instant} and \emph{Viterbi}) improve against the GNN-only KF, resulting in a 10 meter decrease in localization error at the 95$^{th}$ percentile. Our proposed \emph{TGNN} approach decreases the positioning error in challenging scenarios by an additional 13 meters, with less than 1 m error increase at the 50$^{th}$ percentile. This constitutes a meaningful increase in performance for the challenging scenario, where differences of 13 meters can make a significant difference in downstream applications.  %Compared to the GNSS-only KF baseline, this corresponds to a 29\% improvement in HE@95$^{th}$, and a 19\% improvement in HE@50$^{th}$, showcasing the potential of including road network measurements in the KF update. Finally, t
The low standard deviation over 10 random weight initializations displays the robustness of the proposed approach.

\begin{figure}
    \centering
    \vspace{-0.5em}
    \includegraphics[width=0.99\linewidth]{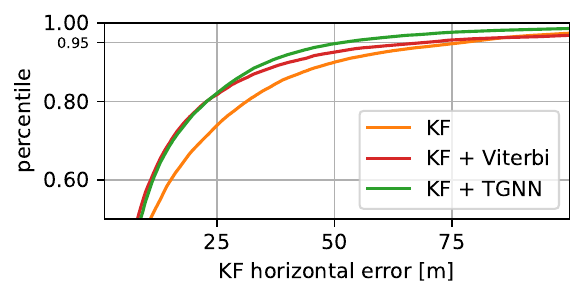}
    \vspace*{-0.3cm}
    \caption{Horizontal Error CDF for different methods.
    }
    \vspace*{-0.2cm}
    \label{fig:cdf}
\end{figure}

We also report the performance for the \emph{Oracle} approach, in which we assume a perfect road network classifier with covariance set to zero (i.e. completely trusting the road network).
Despite the use of the \emph{Oracle}, the positioning error does not go to zero because of three reasons. First, there are mismatches between map information and the actual road network structure. Second, we do not take road width into account and snap to the center of the road, because lane level information is often unavailable. Third, while generally robust, there are occasional failures of the bidirectional Viterbi \emph{Oracle}. 
% Interestingly, we notice that this method does not reduce the positioning error to zero. This is because of three reasons. Firstly, there can be mismatches between the extracted map information and the road network structure on which the ground truth positions were annotated. Roads may be missing on the map or their exact coordinates may have been incorrectly measured. Secondly, we do not take road width into account in our model and always consider the center line as the road measurement, since lane width and count information are often unavailable. Lastly, while generally robust, the bidirectional Viterbi \emph{Oracle} can occasionally fail and select incorrect segments in the most likely trajectories, for example in the case of two parallel roads that connect to the same junctions.
% These insights convince us that we could further improve the end-to-end positioning performance by increasing the fidelity and number of features available for the road network data.
This highlights that performance could be improved by increasing the fidelity and number of features available for the road network data.

In Figure \ref{fig:cdf} we include a visualization of the horizontal error CDFs. We observe that both methods using road network measurements improve over the GNSS-only KF at the median as well as high percentiles, The proposed \emph{TGNN} approach further improves over the Viterbi baseline on the challenging environments above the 90$^{th}$ percentiles.

\subsection{Ablations}
% \db{Changed this section to present tense, let's not mix present and past throughout the paper}
To evaluate the impact of out implementation choices, we conduct four different ablation studies.

\paragraph{Predictor type}
First, we investigate whether both the road selection or covariance prediction components benefit from being modeled through a neural network, by replacing either of them with their non-learned counterparts: \emph{Viterbi} and grid search. As shown in Table \ref{tab_selection_ablations}, the two ablations result in degraded performance, suggesting that both road selection and covariance estimation tasks can be better tackled with a learned model. In particular, differently from grid search, the learned covariance estimation module can dynamically tune the predicted uncertainty at test time, and better adapt to changing test scenarios.

\begin{figure}
    \centering
    \vspace{-0.5em}
    \includegraphics[width=0.99\linewidth]{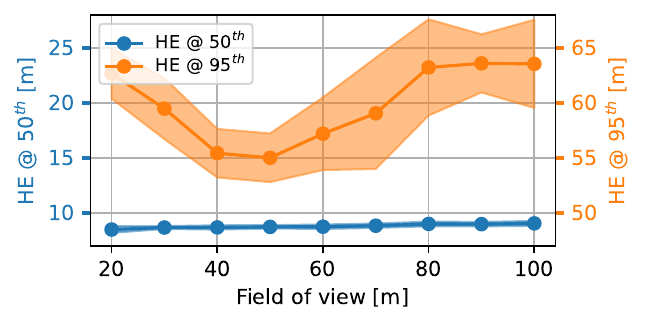}
    \vspace*{-0.2cm}
    \caption{Effect of changing the field of view of the proposed TGNN model on both the 50$^{th}$ and 95$^{th}$ horizontal error percentile. While the error does not change in benign scenarios, the performance in challenging environment significantly decays when choosing too small or large fields of view. 
    % \db{Davide to shrink vertical space, merge the legends and change the y axes to match the orange and blue color} \hg{Yes, sounds good. I don't have access to the raw figure anymore. So we'll have to do this remotely. Maybe in the videocall?}
    }
    \label{fig:fov}
    \vspace{-1em}
\end{figure}

\paragraph{Architecture} Second, we assess the impact of different modules in the \emph{TGNN} architecture by ablating them one by one. The results of these ablations are shown in Table \ref{tab_network_ablations}.
The \emph{GNN} model is obtained by removing the temporal LSTM component. This model cannot capture dynamics that span across multiple time-steps such as changes in vehicle speed and heading, resulting in a 2 meters increase in the positioning error in challenging scenarios. We name \emph{MLP} a simpler architecture variation where the KF state values are simply concatenated to the road features and processed by a sequence of MLPs. Without graph convolutions, information cannot be propagated across road segments in the graph, and the neural network must predict the probability for each segment independently of the others. This results in a further increase in positioning error by 3 meters. We therefore confirm that the neural network must be equipped with both temporal (LSTM) and spatial (GCN) reasoning capabilities to improve its performance. We refer to Appendix \ref{app:netwok_ablations}
for implementation details on the ablated architectures.

\begin{figure*}
    \centering
    \begin{tabular}{cccc}
        \hspace{-1.2em}
         \includegraphics[width=0.25\linewidth] {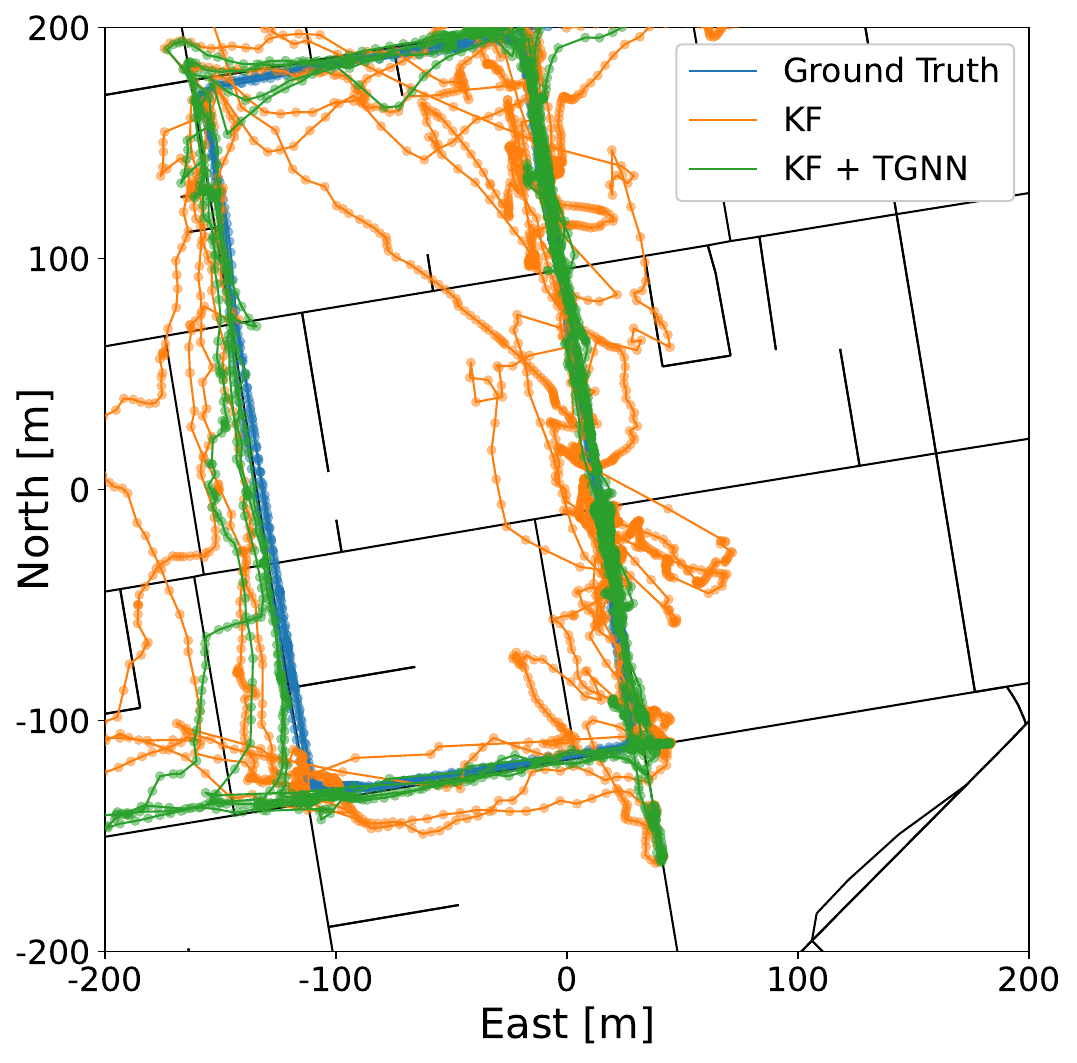} & 
         \hspace{-1.2em}
         \includegraphics[width=0.25\linewidth] {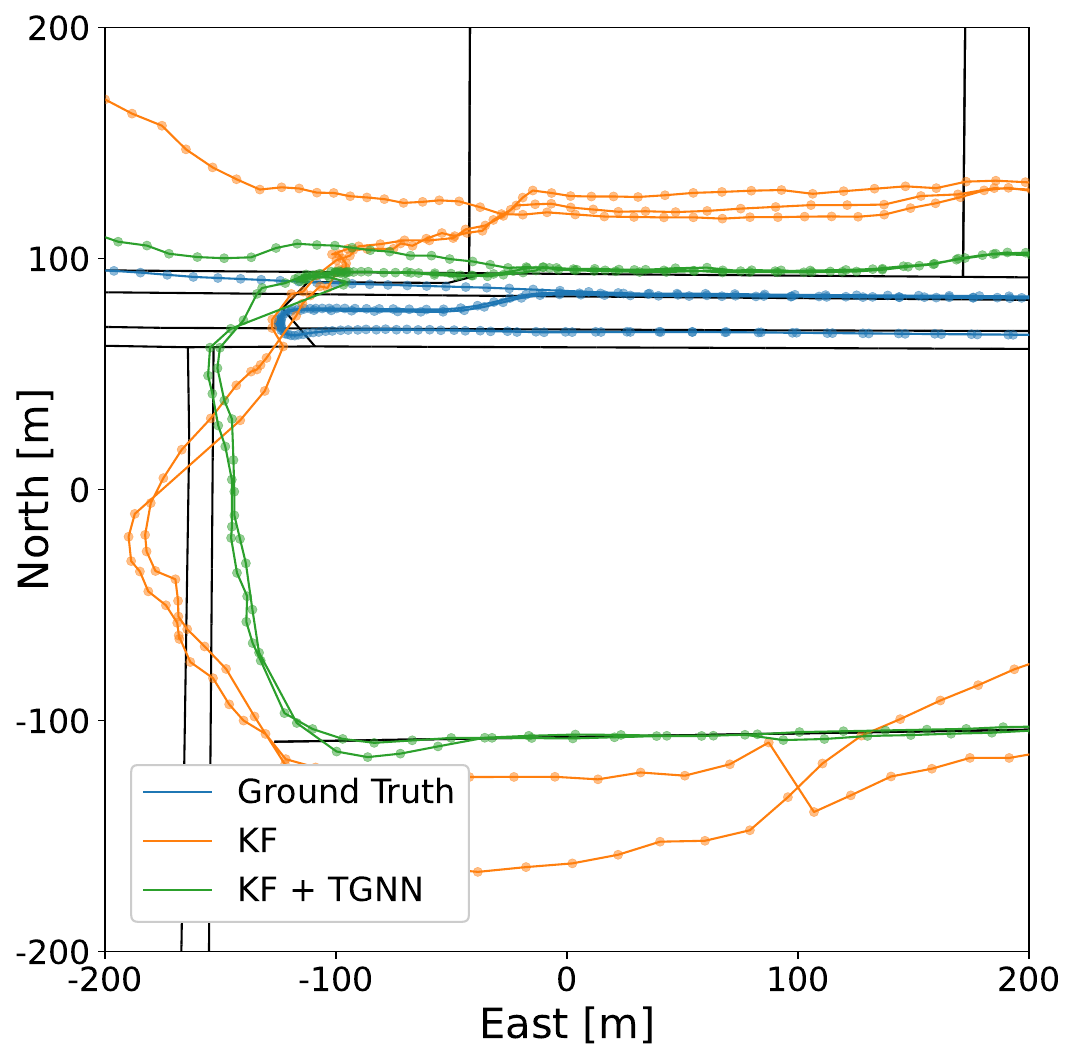} & 
         \hspace{-1.2em}
         \includegraphics[width=0.25\linewidth] {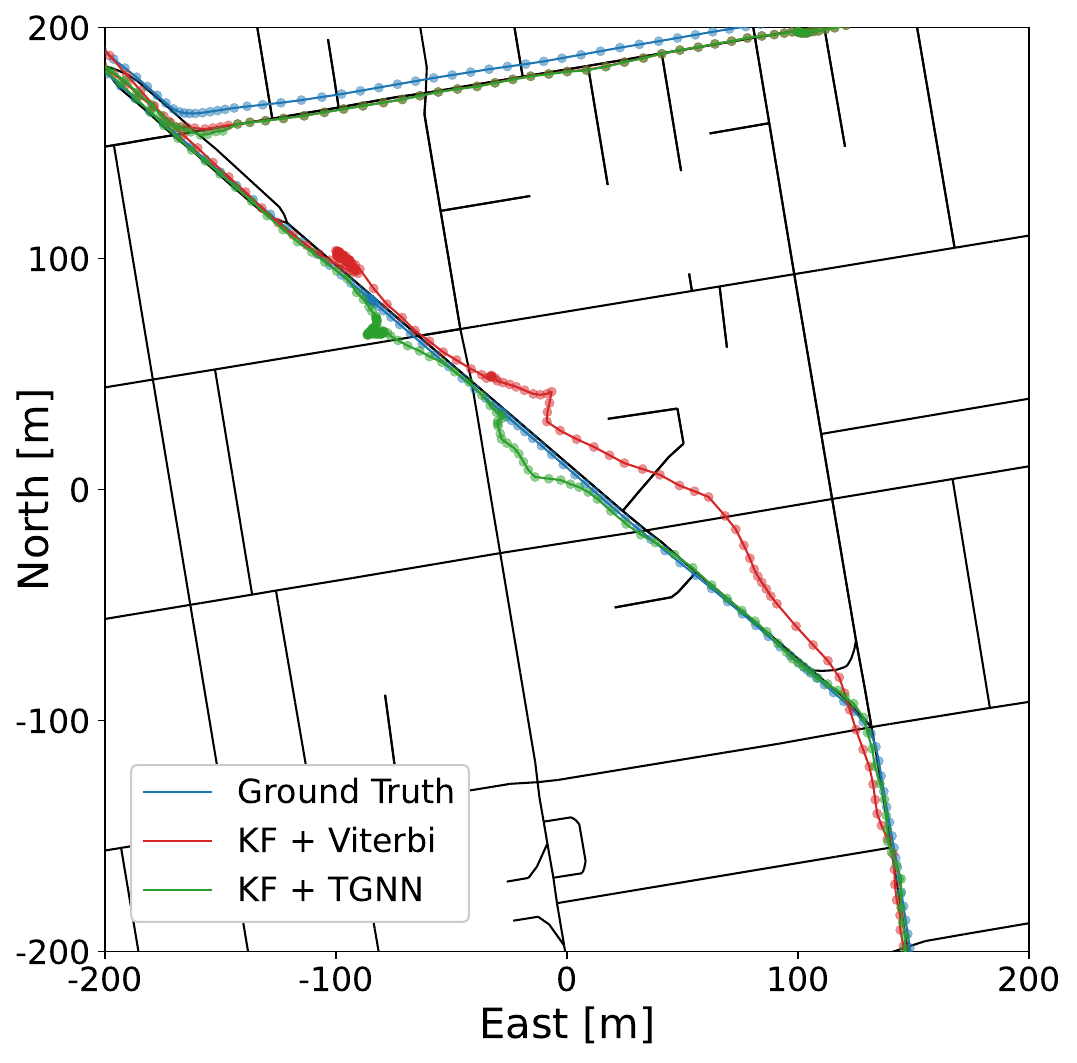} & 
         \hspace{-1.2em}
         \includegraphics[width=0.25\linewidth] {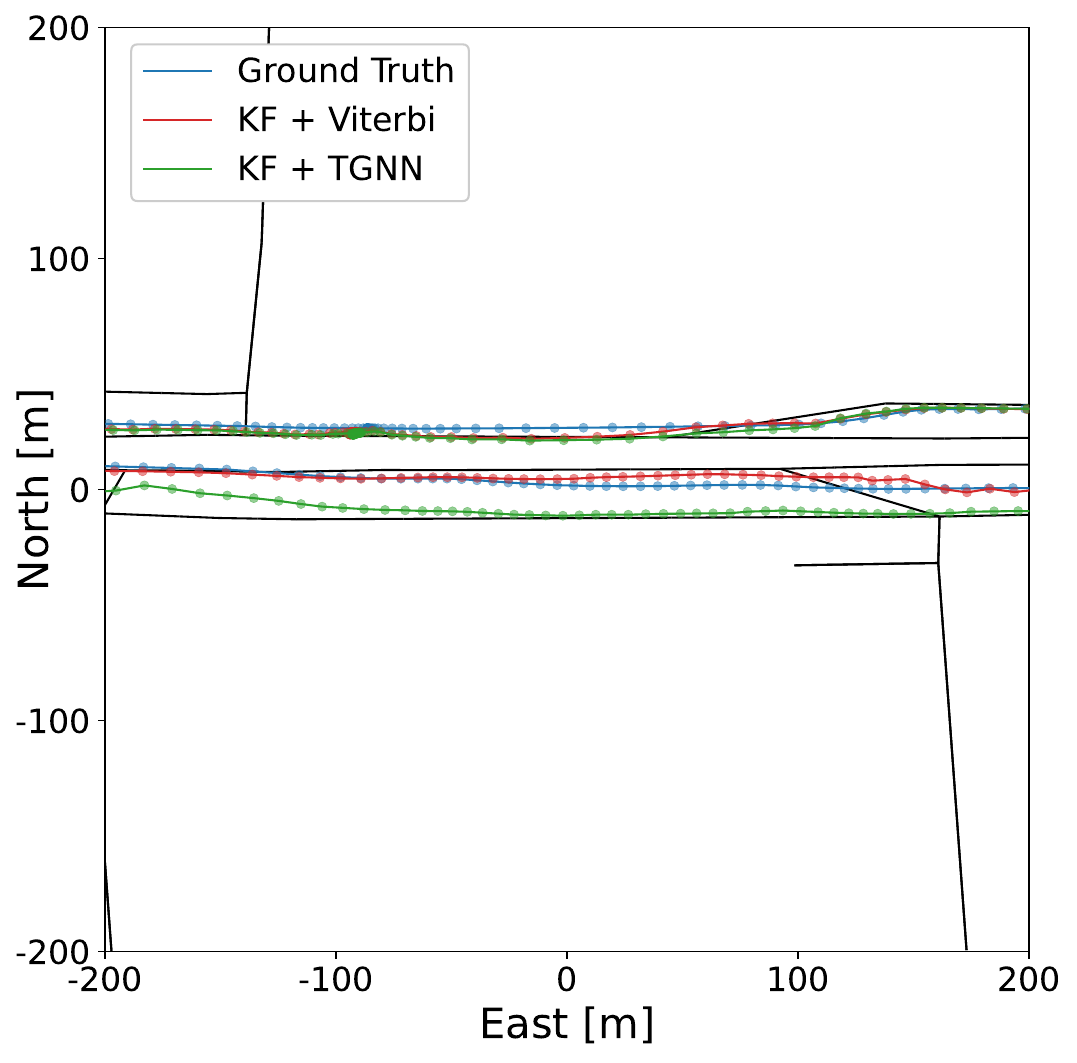}  \\
         \begin{tabular}{c}\scriptsize(a) TGNN {\color{green}{outperforms}} GNSS-only\end{tabular} & 
         \hspace{-1.2em}
         \begin{tabular}{c}\scriptsize(b) TGNN {\color{red}underperforms} GNSS-only\end{tabular} & 
         \hspace{-1.2em}
         \begin{tabular}{c}\scriptsize(c) TGNN {\color{green}{outperforms}} Viterbi\end{tabular} & 
         \hspace{-1.2em}
         \begin{tabular}{c}\scriptsize(d) TGNN {\color{red}underperforms} Viterbi\end{tabular} 
    \end{tabular}
    \vspace{-0.3em}
    \caption{Qualitative results for the proposed TGNN model compared against the Viterbi baseline, zoomed in to interesting parts of the drive. Full images can be found in supplementary material Appendix E %\ref{app:qualitative}
    .}
    \vspace{-1.2em}
    \label{fig:qualitative}
\end{figure*}

% \begin{figure*}
%     \centering
%     \begin{tabular}{cc}
%          \includegraphics[width=0.4\linewidth]{qualitative/paper/best_for_LSTM_over_viterbi_0.pdf} & 
%          \includegraphics[width=0.4\linewidth]{qualitative/paper/best_for_viterbi_over_LSTM_0.pdf} \\
%          \begin{tabular}{c}\footnotesize\textbf{A}: Drive where TGNN has the highest \\ \footnotesize improvement over Viterbi.\end{tabular}  & 
%          \begin{tabular}{c}\footnotesize\textbf{B}: Drive where TGNN has the lowest \\ \footnotesize improvement over Viterbi.\end{tabular}
%     \end{tabular}
%     \caption{Qualitative results for the proposed TGNN model versus Viterbi, zoomed in to interesting parts of the drive. Full images can be found in Appendix \ref{app:qualitative}. \hg{The legend is missing from these figures. Blue is GT, orange is Viterbi, geen is TGNN} \db{we might want to add a third figure, maybe with "average" improvements, to fill the page horizontally and also take less vertical space (existing images shrink)}}
%     \label{fig:qualitative}
% \end{figure*}

\paragraph{Input features} Third, we test the effect of ablating several road input features, as shown in Table \ref{tab_feature_ablation}. The costs assigned to each road using Equations \ref{eq:emission_position} and \ref{eq:emission_heading} are the most impactful, with additional features such as road type and maximum driving speed being also important. However, the best results are obtained when using all the available features (listed in Appendix \ref{app:dataset}). Interestingly, a model with only the \emph{Viterbi} prior input features, which describe the probability of the roads assigned in the previous time step, can also improve over the standard KF. This could be the case in situations where the GNSS measurements are noisy, while the road selection task is trivial (only one road is present in the receptive field) and can therefore improve the KF state. 
% \db{Hans check if this explanation can make sense, and correct otherwise. I wanted to say something about this as the reviewers might find it weird that we can improve without input features.} \hg{Yes this makes sense! Good to point it out already, beacuse indeed this is something reviewers might trip on.}

\paragraph{Field of view} Finally, we experiment with changing the field of view for our method, which defines the set of road segment candidates for selection based on a radius around the user position estimate. We visualize in Figure \ref{fig:fov} how changing this parameter impacts the positioning error. The performance at the 95th percentile degrades quickly when the field of view is too large or too small. In the former case, this is likely because the correct road to select shifts out of focus more frequently, while in the latter case, the number of candidate road segments increases significantly, which might make the classification task more challenging for the neural network. The field of view is fixed to 50 meters in all other experiments.

\subsection{Qualitative Results}
We include a qualitative analysis to visualize and discuss patterns in the model behavior. Figure~\ref{fig:qualitative} zooms in on interesting scenarios, while in Appendix~\ref{app:fullfigures}
we include the qualitative results for the full drives.
We first compare the proposed KF + \emph{TGNN} against a simple KF using only GNSS measurements. In Figure \ref{fig:qualitative}a we show a drive where \emph{TGNN} significantly outperforms the GNSS-only baseline. In this challenging urban scenario the GNSS signal is very noisy, which results in unstable and incorrect predictions for the simple KF baseline. By using the road network measurements, the \emph{TGNN} can compare the noisy GNSS measurements with the prior knowledge about the road structure, and significantly reduce the localization error. In Figure \ref{fig:qualitative}b we show \emph{the only drive} in which the \emph{TGNN} underperforms the simple KF baseline. We notice how the large GNSS noise causes both solutions to drift away from the ground-truth trajectory. However, the \emph{TGNN} solution confidently snaps to a road segment well-aligned to the noisy GNSS measurements but disconnected to the correct segment, which prevents the KF from recovering its state over time. We hypothesize that this performance drift could be significantly limited by incorporating an IMU component in the KF, as measuring the vehicle heading could prevent selecting road segments with incorrect heading.

Next, we compare the \emph{TGNN} against the \emph{Viterbi} baseline to process road network measurements. In Figure \ref{fig:qualitative}c we show an example in which the proposed method outperforms the baseline by confidently following the correct road segment. Figure \ref{fig:qualitative}d describes a road network with a few parallel roads with multiple lanes each. In this case, the \emph{TGNN} incorrectly selects a road segment with a consistent offset with respect to the ground truth trajectory. In addition to using IMU measurements, we believe that a lane-level representation of the network might reduce these errors, as the Kalman Filter would not be forced to follow one of the two road center lanes, but could instead select an intermediate lane segment better matching the current observations.

% The two parts of the drives shown in Fig. \ref{fig:qualitative} are in very challenging dense urban environments. This is reflected in the large positioning errors made by the methods. In the case of the lowest improvement especially, Fig. \ref{fig:qualitative}B shows that both methods completely diverge from the correct solution. In such cases, it is arguable whether road network information could prevent such KF divergence, or whether other techniques, such as measurement filtering and correction, would be more suitable. 

\section{Conclusions}
In conclusion, this paper presents a novel method for enhancing GNSS-based vehicle positioning by integrating road network information into a KF using a Temporal Graph Neural Network. To the best of our knowledge, this is the first approach employing a neural network to process the road network data in a positioning system. Our approach significantly reduces horizontal positioning errors in real-world challenging scenarios, demonstrating the effectiveness of our data-driven solution compared to other methods.

Future work could explore using a more fine-grained representation of the road network, such as including lane-level information, or using a visual representation of the map. Another novel research direction would be learning to filter or correct the GNSS measurements based on the road network structure before using them in the KF update step.


\ifcameraready
\section*{Acknowledgments}

We would like to thank Maximilian Arnold for providing helpful feedback on this manuscript.
\fi

\section*{Impact Statement}

This paper presents work whose goal is to advance the field of Machine Learning for GNSS positioning. Positioning systems can be embedded in a variety of real-world applications. Positioning systems deployed for critical applications such as autonomous driving should be thoroughly tested and integrated with strong safety measures, both software and hardware, to ensure the highest standard of safety for the consumer. 

There may be other potential societal consequences of our work, none which we feel must be specifically highlighted here.

\bibliographystyle{icml2025}
\bibliography{main}

\newpage
\appendix
\onecolumn

%%%%%%%%%%%%%%%%%%%%%%%%%%%%%%%%%%%%%%%%%%%%%%%%%%%%%%%%%%%%%%%%%%%%%%%%%%%%
\section{Deriving z, H and V}
\label{app:equations}
For Equation~2%\ref{eq:kf_update}
, we need to define the measurement matrix $\rmH$, the observation vector $\rvz$ and the covariance matrix $\rmV$. The measurement matrix consists of two parts. First we select from the KF state-space vector $\rvx_{\operatorname{GNSS}}^+$ only the current user position $\rvx_{\text{pos}}$ in terms of $x$ and $y$ components, which will be updated with the road network measurement. This is obtained through the selection matrix $\rmH_{\text{select}}$. Secondly, we rotate $\rvx_{\text{pos}}$ into a perpendicular and  parallel component with respect to the selected road using the clockwise rotation matrix $\rmH_{\text{rot}}$:
\begin{equation}
    \rmH = \rmH_{\text{rot}} \rmH_{\text{select}} = \begin{bmatrix}
   \cos{\theta} &
   \sin{\theta} \\
   -\sin{\theta} &
    \cos{\theta} 
   \end{bmatrix} \begin{bmatrix}
   1 &
   0 & \dots & 0 \\
   0 &
   1 & \dots & 0
   \end{bmatrix}
\end{equation}

where $\theta$ is the heading of the selected road $r^*$. Multiplying $\rmH$ with $\rvx_{\operatorname{GNSS}}^+$ thus results in: %\db{proposing a more compact refactoring since we already defined $\rmH_{\text{rot}}$ above, and also to try and make clearer where $\rvx_{\text{pos}}$ comes from}
\begin{align}
    \rmH\rvx_{\operatorname{GNSS}}^+ &= 
 \rmH_{\text{rot}} \rvx_{\text{pos}} =  
   %  \begin{bmatrix}
   % \cos{\theta} &
   % -\sin{\theta} \\
   % \sin{\theta} &
   %  \cos{\theta} 
   % \end{bmatrix} \rvx_{\text{pos}} =  
   \begin{bmatrix}
   x_{\parallel} \\
   x_{\perp} 
   \end{bmatrix},
\end{align}
% \db{The equation concludes by finding $\rmH\rvx$, even if we are looking for the $\rmH$ only.} \hg{split into two equations now}\db{thanks, much clearer now.}
where $\rvx_{\parallel}$ is the parallel and $\rvx_{\perp}$ the perpendicular component of the user location projected on the selected road. 
% \db{If I understand correctly, these are not the components of the observation, but the components of the user location projected to observation (road network) coordinate space. Can you confirm and if so clarify in text?} \hg{Correct! I have changed the text.}

The mean of the observation is a vector of size two, also consisting of a perpendicular and parallel component. The mean can be found by rotating the center point of the road $\rvr^*_{\text{pos}}$ in the clockwise direction as: %\db{what do you mean here? What is the 'other direction'?} \hg{The user is rotated counterclockwise, and the road is rotated clockwise} \db{got it! should we add an intermediate step in the equation to clarify that we are inverting the rotation from $\rmH_{\text{rot}}$? see proposal in blue}
\begin{equation}
    \rvz  = 
    \rmH_{\text{rot}} \rvr^*_{\text{pos}} = 
   \begin{bmatrix}
   z_{\parallel} \\
   z_{\perp} 
   \end{bmatrix}.
   \label{eq:z}
\end{equation}

Finally, we soft-threshold $z_{\parallel}$ in order to project the vehicle onto the closest point on the road, for cases where the orthogonal projection would fall outside the road segment extreme points. This can be done by soft-thresholding $z_{\parallel}$ using the length of the road segment $r^*_l$:
\begin{equation}
    z_{\parallel}^* = \text{sign}\left(z_{\parallel}\right) \max\left(|z_{\parallel}| - \frac{1}{2}r^*_l, 0\right).
\end{equation}

The covariance of the road observation can then be simply expressed with two variables along the diagonal as:
\begin{equation}
\label{eq:road_covariance}
    \rmV = \begin{bmatrix}
   \sigma_{\parallel}^2 &
   0 \\
   0 &
   \sigma_{\perp}^2
   \end{bmatrix},
\end{equation}
where $\sigma_{\parallel}^2$ is the variance of the road observation parallel to the road and $\sigma_{\perp}^2$ is the variance perpendicular to the road. These parameters can be tuned on some calibration data or predicted with a neural network, as we will discuss later in this section. 

%%%%%%%%%%%%%%%%%%%%%%%%%%%%%%%%%%%%%%%%%%%%%%%%%%%%%%%%%%%%%%%%%%%%%%%%%%%%
\section{Dataset}
\label{app:dataset}
We provide in this section additional details regarding the processing of GNSS and Road Network data for our experiments.

\paragraph{GNSS data}
User location and time was obtained from geodetic-survey grade receiver with a mobile phone form factor antenna and an Inertial Measurement Unit. The pseudo-ranges were measured at a 1Hz frequency using the position data and the satellite constellation orbital data, with the receiver’s clock bias measured by a stable external reference clock source. Ionospheric and tropospheric delays were corrected using Klobuchar \citep{klobuchar1987ionospheric} and Saastamoinen \citep{bevis1994tropospheric} models, respectively. Additionally, the Sagnac effect due to the Earth's rotation was corrected for \citep{bidikar2016sagnac}. The time biases between the different GNSS constellations were estimated using a preliminary weighted least squares (WLS) localization and removed, allowing the model to use pseudorange measurements from all constellations. To attain this preliminary WLS, we can make use of a localization acquired by a cellular network or an externally injected position available on the device.

\paragraph{Road network data}
We use the OSMnx package \citep{osmnx} to extract the road network for a bounding box around each GNSS drive route. We only retain the driving road network by specifying the OSM tag \enquote{highway}\footnote{\url{https://wiki.openstreetmap.org/wiki/Key:highway}}, which removes other structures such as cycleways and footways.
The extracted road segments are of varying lengths, with some spanning over 500 meters long while others being just 1 meter. To make the segments length more consistent and easier to process by the various methods, we divide all roads longer than 25 meters into equal segments up to 25 meters in length. As a last step, we convert the graph to its dual, such that all road segments become nodes, and all intersections become edges for the purpose of the graph convolution operations (see Figure 2%\ref{fig:viterbi}
).

%%%%%%%%%%%%%%%%%%%%%%%%%%%%%%%%%%%%%%%%%%%%%%%%%%%%%%%%%%%%%%%%%%%%%%%%%%%%
\subsection{Input features}
% \db{this paragraph is currently only describing the road features, but we also need a sentence about the vehicle input features. We can add for clarification that we are talking about $\rmR$ and $\rvx$. If the paragraph becomes too long, we can switch to subsections.}
We here describe the most useful road and vehicle input features, denotes by $\rmR$ and $\rvx$, respectively. 

\paragraph{Road Features} 
The most useful road features were found to be: \emph{Distances}, \emph{Road Type}, \emph{Max Speed}, \emph{Road Heading}, \emph{Oneway}, and \emph{Viterbi Prior}. \emph{Distances} consists of the Euclidean distance calculated using Equation~3 %\ref{eq:emission_position}
and the angular distance calculated using Equation~4%\ref{eq:emission_heading}
. The \emph{Road Type} feature is a one-hot encoding of the road type as provided by OpenStreetMap. While the \enquote{highway} tag covers 27 different types of roads, not all of them are relevant to our applications (e.g., via\_ferrata). Therefore, we only one-hot encoded the following types: `motorway', `motorway\_link', `trunk', `trunk\_link', `primary', `primary\_link', `secondary', `secondary\_link', `tertiary', `tertiary\_link', `unclassified'\footnote{From OpenStreetMap: ``The word `unclassified' is a historical artefact of the UK road system and does not mean that the classification is unknown"}, `residential', `living\_street', `service'. All other types defaulted to a separate value in the one-hot vector. We provide \emph{Max Speed} as the maximum driving speed in meters per second. To the above input features, we add an additional Viterbi prior feature. \emph{Road Heading} describes the road heading in polar coordinates. To that end, we provide the sine and cosine of the azimuth of the road, which automatically encodes the modulo nature of the azimuth. \emph{Oneway} is a simple boolean feature that encodes whether the road is oneway (1) or bidirectional (0). Lastly, to mimic the Viterbi algorithm's capability to track state probabilities over time, we append the probabilities assigned to each road to the road features $\mathbf{R}$ of the next time step. Furthermore, we propagate these probabilities to the $k$-hop neighbors (following Equation~6%\ref{eq:transition}
) and take the maximum. For each $k \in {1, \dots, K}$, this maximum neighboring probability is added as a feature to $\mathbf{R}$ for the next time step. 

\paragraph{Vehicle Features} 
The most useful vehicle features were found to be: \emph{Vehicle Heading} and \emph{Position Uncertainty}. \emph{Vehicle Heading} desribes the azimuth of the vehicle's heading in the same way as the \emph{Road Heading} feature, i.e., in terms of the sine and cosine of the azimuth. Furthermore, we also provide the magnitude of the estimated speed in meters per second. This allows TGNN and to compare the \emph{Vehicle Heading} with the \emph{Road Heading} and \emph{Max Speed} features of each road in a straight-forward manner. The \emph{Position Uncertainty} feature encodes the covariance matrix of the KF in terms of the uncertainty about East and North. That is, we provide $\rmV_{\text{pos}}$ in terms of the three scalars $\sigma_{xx}^2, \sigma_{xy}^2, \sigma_{yy}^2$. Note that providing the other covariance $\sigma_{yx}^2$ would be redundant with $\sigma_{xy}^2$.

%%%%%%%%%%%%%%%%%%%%%%%%%%%%%%%%%%%%%%%%%%%%%%%%%%%%%%%%%%%%%%%%%%%%%%%%%%%%
\section{Hyper-parameters}
\label{app:hyper}
% \db{not sure if we should mention doing a search on the calibration set or not.} \hg{This reads as though I searched for all parameters on the calibration set, but I only searched for the two values of sigma.} \db{agreed and removed the part on the search. It remains that a reviewer might ask why we chose this parameter. Not sure if we can say anything to prevent that comment.}
We report in this section all implementation details and hyper-parameter choices to facilitate reproducing our experimental results. The K-hop distance considered in the Viterbi algorithm is set to $K=2$ to increase robustness in high-speed scenarios. The cost weighting for the emission probability is set to $\beta = 0.01$, such that a position difference of 100 meters is weighted equally as a difference in heading of 90 degrees. We found this trade-off to show the best results in preliminary experiments with the \emph{Viterbi} baseline, and fixed the hyper-parameter for all our experiments.

% \db{if we need to carve out space, consider moving this part to appendix. it might be a bit complicated to grasp for the reader.} \aj{Is $z_\perp$ always 0? I did not get why we need to rotate both x and r, isn't one enough? BTW, if we are voting, I would not remove the equations and discussions above from here.} \hg{No, $z_\perp$ is learned or searched and often not equal to zero. Of course, there is nothing stopping the learning algorithm or search algorithm to set $z_\perp$ to zero. But that only happened in practice for the genie, as that was the only case for which it was the best choice to `trust' the road measurement completely. As for the rotation, I double checked my math and indeed in the original notation in the paper there was the mistake of rotating them both in opposite directions. This is not the case, both should be rotated clockwise, updated this now. You do need to rotate both x and z though, so that they are in the same coordinate space. This can especially be seen when the KF state is centered such that its x-y position is zero. (As is the case in our implementation)}

To choose the road measurement variances in methods without uncertainty prediction, we perform a grid search over the values: $\sigma_{\perp}^2 \in \{0, 1, 2, 3, \dots, 10 \}$ and $\sigma_{\parallel}^2 \in \{0, 1, 2, 3, \dots, 10, 100, 200, 300, \dots, 1000, \infty\}$. We calculate the horizontal error at the 95th error percentile for each combination on the training set. The combination with the lowest horizontal error at the 95th percentile is subsequently selected to be used on the hold-out test set. This strategy has the limitation that for all fixes under consideration, the same variance terms are used.

The TGNN architecture consists of $L=4$ blocks and all hidden sizes are set to $32$. We use SiLU activation \citep{elfwing2018sigmoid} after each layer paired with batch normalization \citep{ioffe2015batch}. When learning the road variances with TGNN, the cost weighting for the CE + MSE loss is set to $\lambda=0.01$ such that 100 meters of position distance becomes unit cost. We train TGNN with a batch size of 8 over 5000 iterations. We use the Adam optimizer \citep{kingma2014adam} with a learning rate of 0.001 and a weight decay of 0.001. % \hg{exponential decay rate was not used}

\paragraph{Computational Costs} The TGNN model has less than 50k parameters and its compute cost amounts to 1.7 MFLOPs, which is low enough to achieve real-time inference performance even on efficient embedded processors.

%%%%%%%%%%%%%%%%%%%%%%%%%%%%%%%%%%%%%%%%%%%%%%%%%%%%%%%%%%%%%%%%%%%%%%%%%%%%
\section{Neural Network Ablations}
\label{app:netwok_ablations}

% \begin{figure}[t!]
%     \centering
%     \includegraphics[width=0.99\linewidth]{figures/gnn.png}    \caption{Architecture for the \emph{GNN} ablation.}
%     \label{fig:gnn}
    
%     \vspace{-0.5em}
% \end{figure}
% \begin{figure}[t!]
%     \centering
%     \includegraphics[width=0.99\linewidth]{figures/mlp.png}    \caption{Architecture for the \emph{MLP} ablation.}
%     \label{fig:mlp}
    
%     \vspace{-0.5em}
% \end{figure}

\begin{figure*}[tbh!]
    \centering
    \begin{minipage}{0.45\textwidth}
        \centering
        \includegraphics[width=\columnwidth]{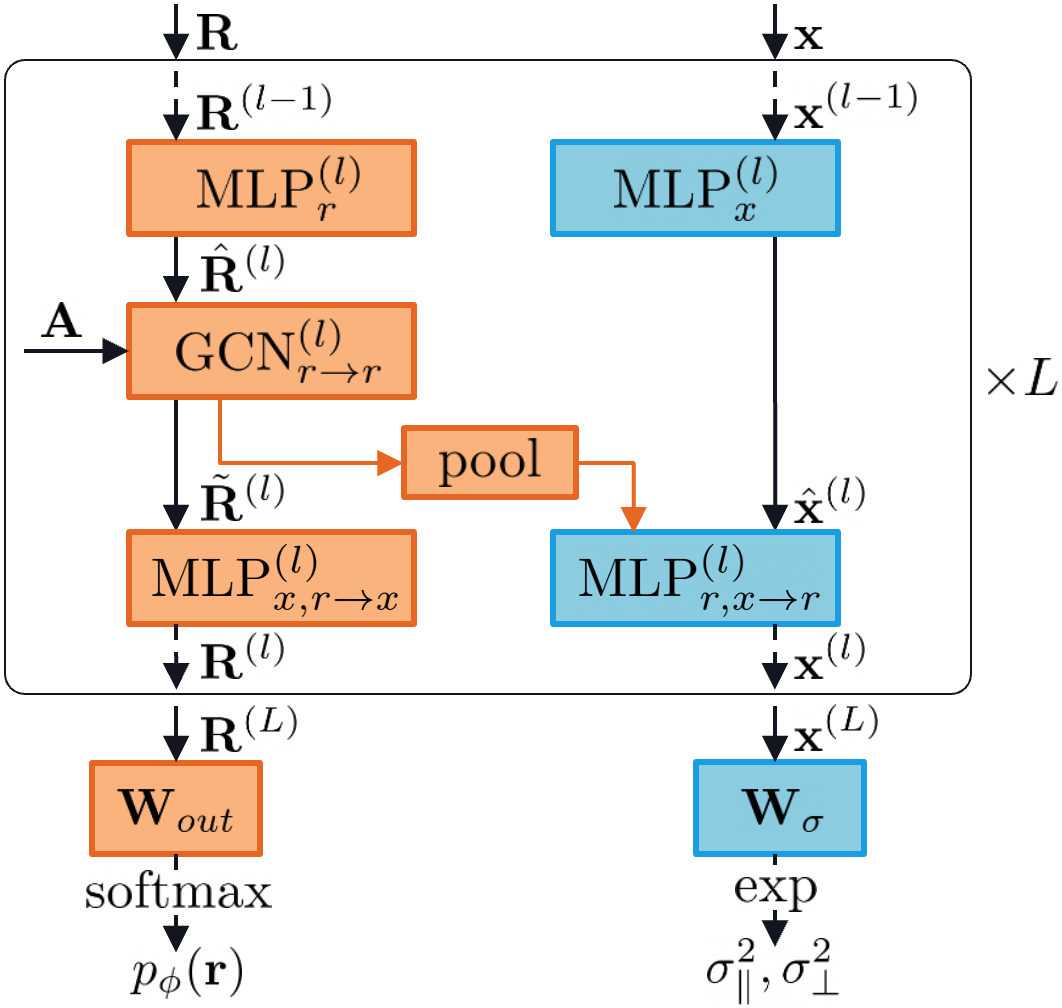}
        \label{fig:gnn}
        {\small ~\\ (a) \emph{GNN} ablation.}\\
    \end{minipage}
    \begin{minipage}{0.03\textwidth}
    ~    
    \end{minipage}
    \begin{minipage}{0.45\textwidth}
        ~\\~\\~\\~\\~\\
        \centering
        \includegraphics[width=\columnwidth]{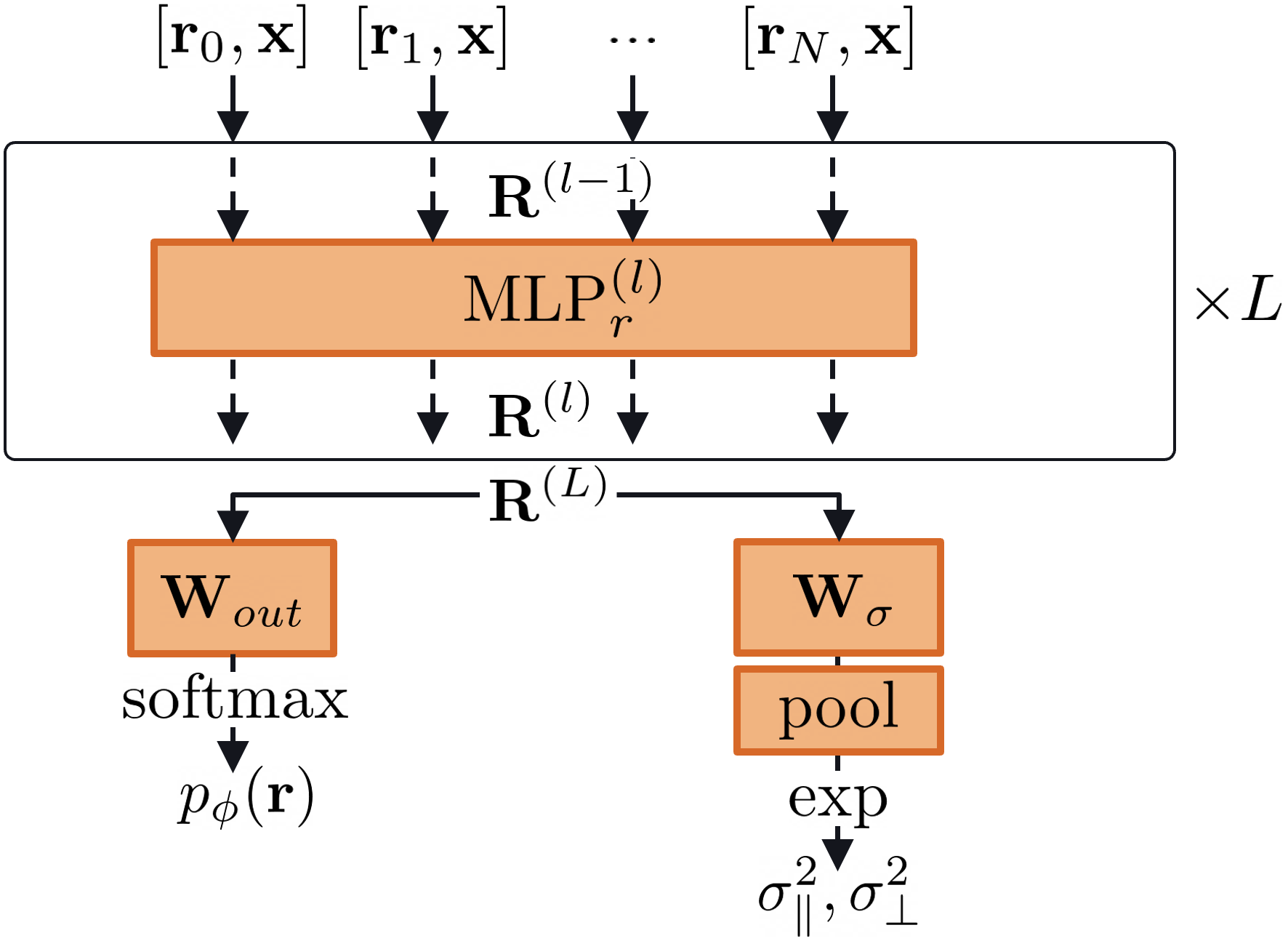}
        \label{fig:mlp}
        {\small ~\\ (b) \emph{MLP} ablation.}\\
    \end{minipage}
    \caption{Architecture for the \emph{GNN} and \emph{MLP} ablations.}
    
    \vspace{-0.5em}
\end{figure*}

We provide diagrams describing the neural network for the ablations of the \emph{TGNN} architecture. The \emph{GNN} model (Figure~7%\ref{fig:gnn}
) is obtained by removing the temporal LSTM component. This model cannot capture dynamics that span across multiple time-steps such as changes in vehicle speed and heading. We name \emph{MLP} (Figure~8%\ref{fig:mlp}
) a simpler architecture variation where the KF state values are simply concatenated to the road features and processed by a sequence of MLPs. Without graph convolutions, information cannot be propagated across road segments in the graph, and the neural network must predict the probability for each segment independently of the others.

%%%%%%%%%%%%%%%%%%%%%%%%%%%%%%%%%%%%%%%%%%%%%%%%%%%%%%%%%%%%%%%%%%%%%%%%%%%%
\section{Full Qualitative Figures}
\label{app:fullfigures}
We include in Figures 9 %\ref{full1}
to 12 %\ref{full4} 
full-size qualitative visualizations of model performance in different drives. Figures~9 %\ref{full1} 
and 10 %\ref{full2} 
compare the proposed KF + TGNN approach against a simple KF using only GNSS measurements. Figures 11 %\ref{full3} 
and 12 %\ref{full4} 
compare KF + TGNN against a the KF + Viterbi baseline to process road network measurements.
\label{app:qualitative}
\begin{figure*}[hbt!]
    \centering
    \includegraphics[width=0.99\linewidth]{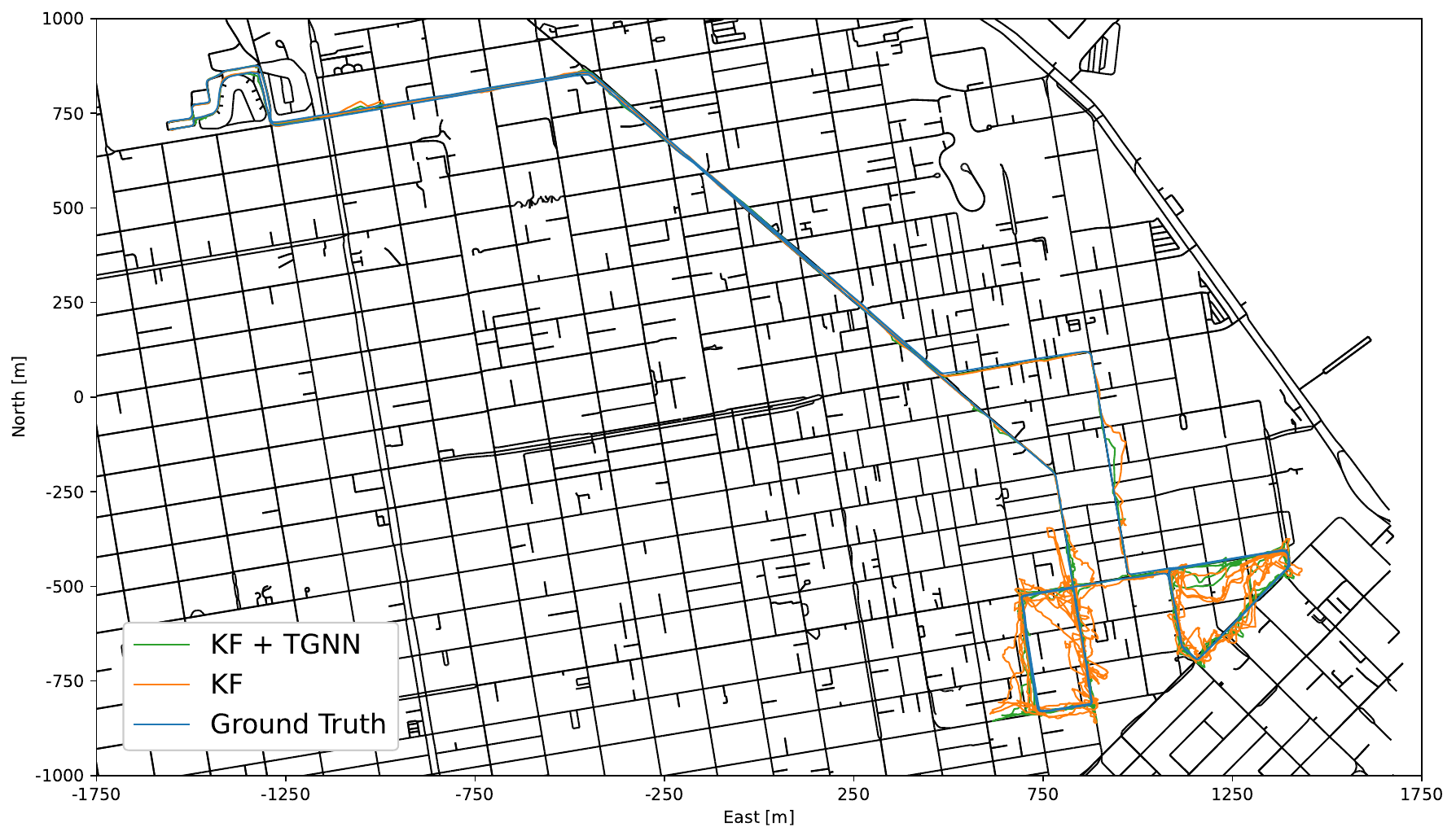}
    \caption{Full drive where TGNN {\color{green}{outperforms}} GNSS-only.}
    \label{full1}
\end{figure*}
\begin{figure*}[hbt!]
    \centering
    \includegraphics[width=0.99\linewidth]{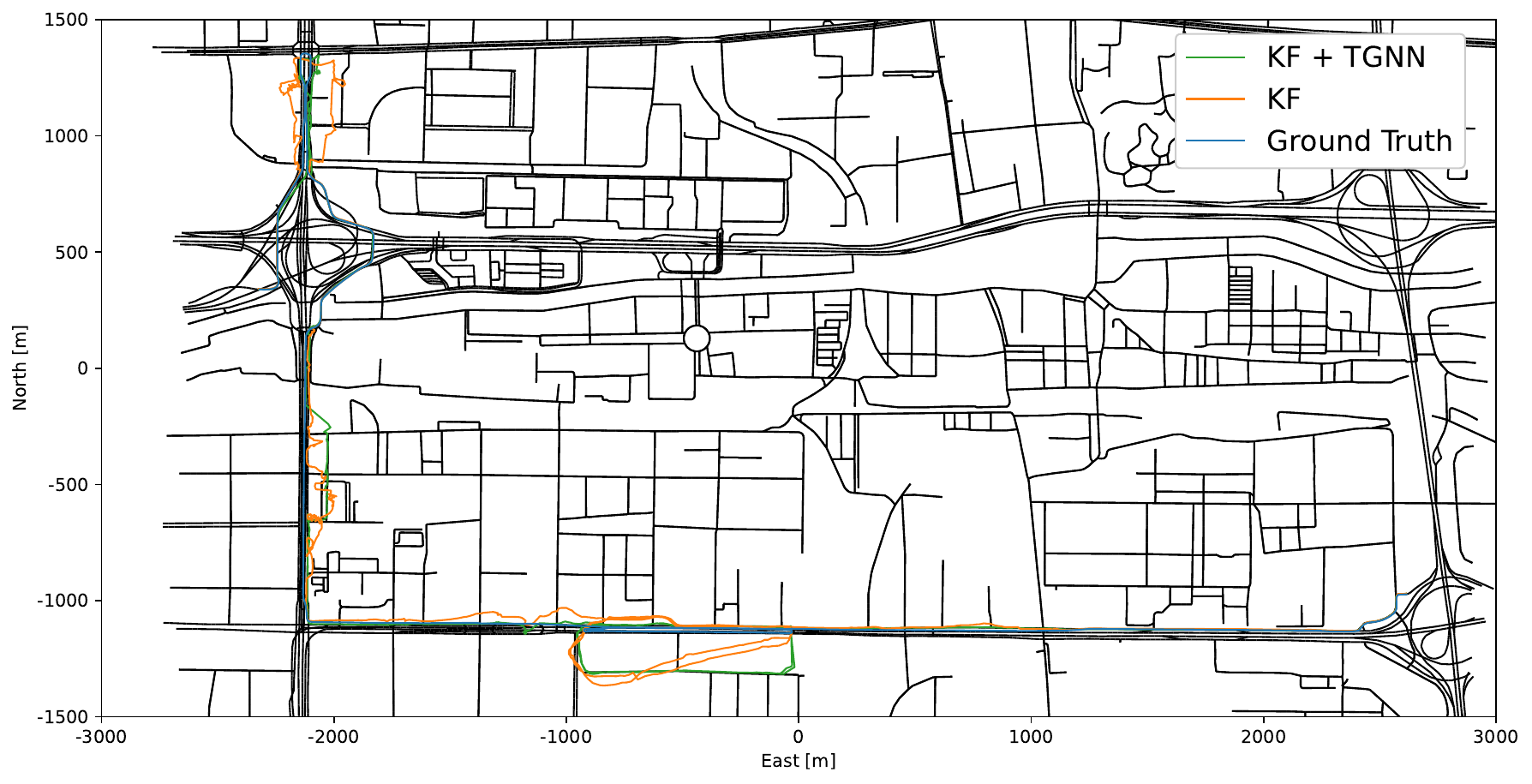}
    \caption{Full drive where TGNN {\color{red}{underperforms}} GNSS-only.} 
    \label{full2}
\end{figure*}
\begin{figure*}[hbt!]
    \centering
    \includegraphics[width=0.99\linewidth]{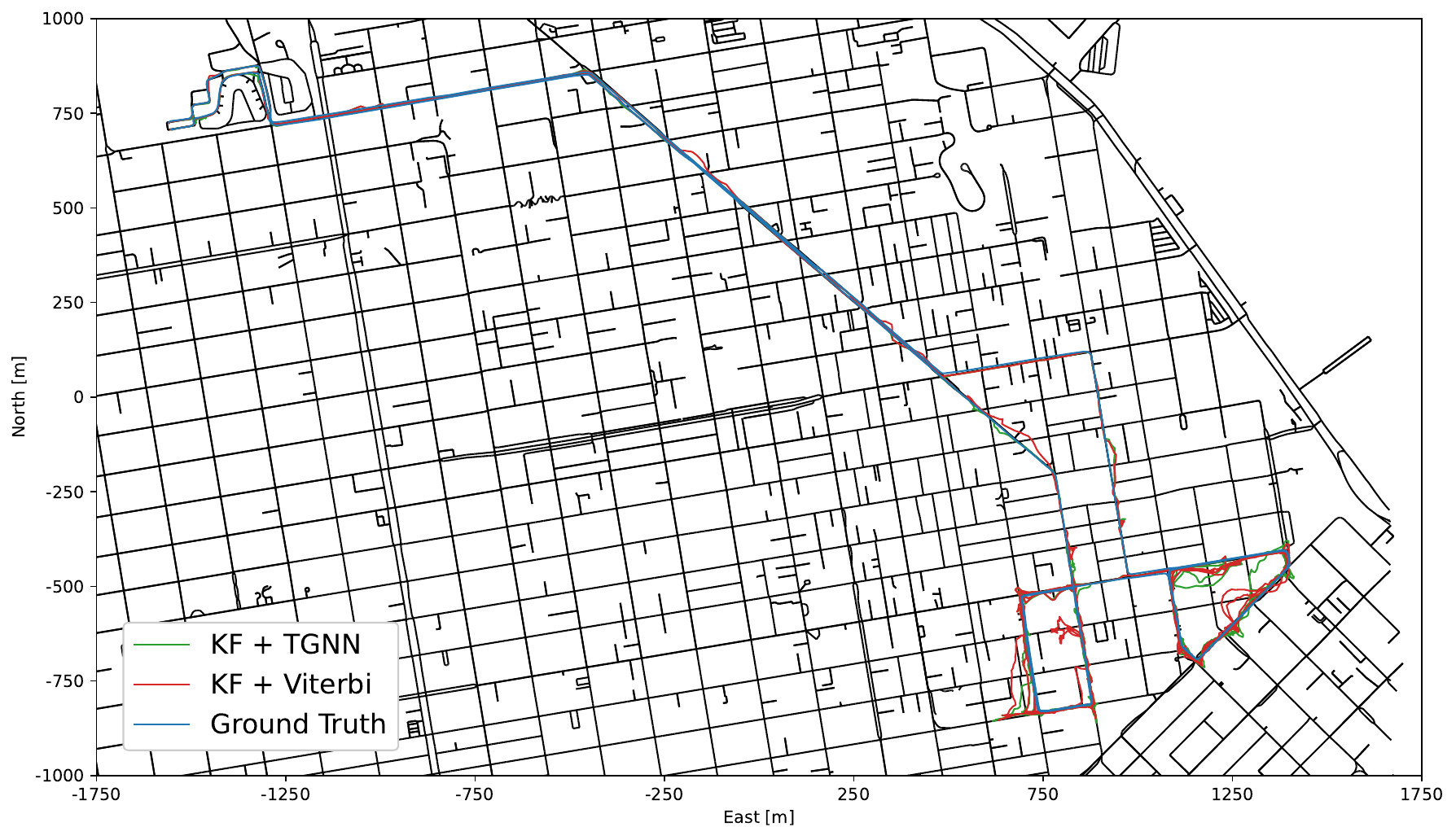}
    \caption{Full drive where TGNN {\color{green}{outperforms}} Viterbi.} 
    \label{full3}
\end{figure*}
\begin{figure*}[hbt!]
    \centering
    \includegraphics[width=0.99\linewidth]{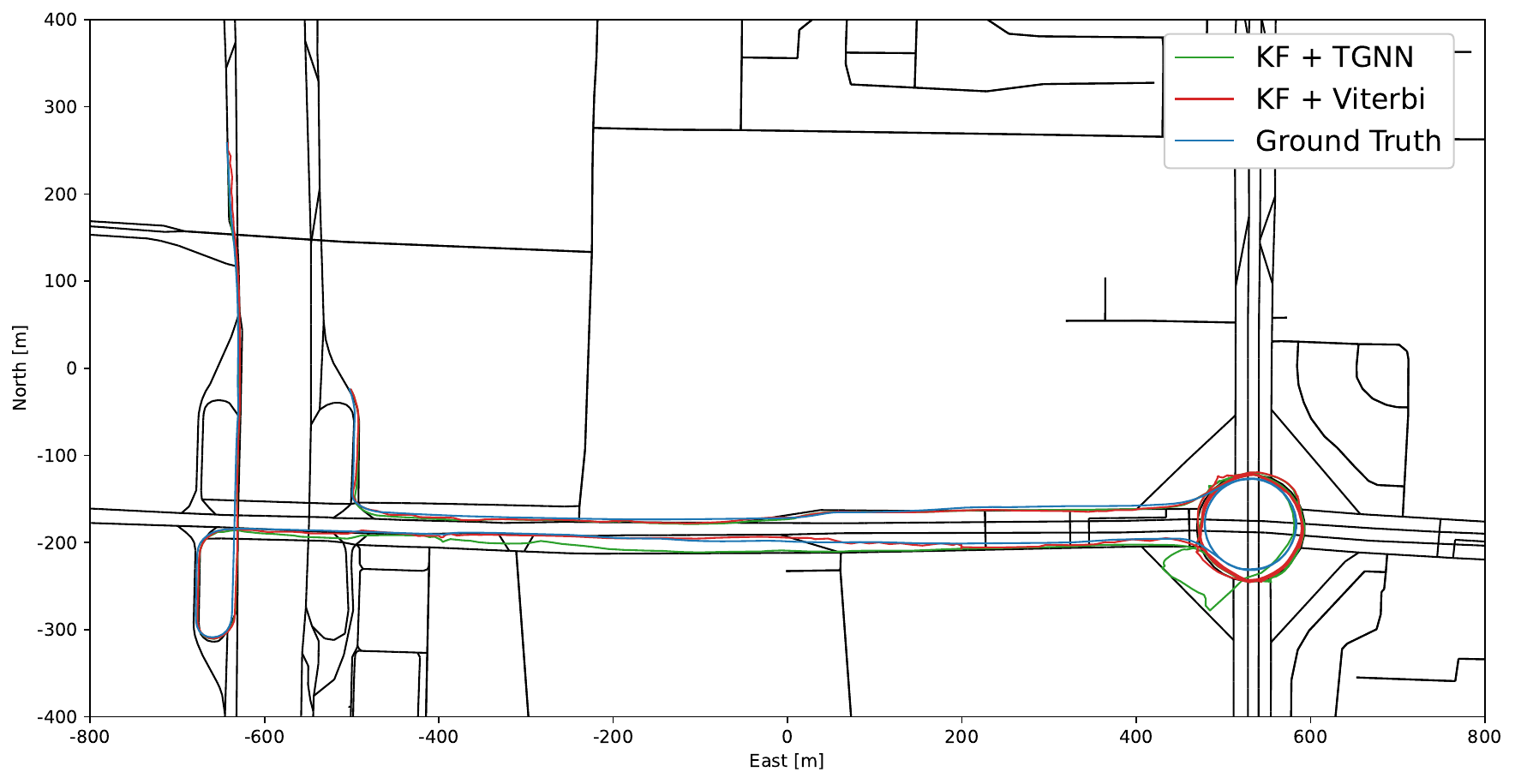}
    \caption{Full drive where TGNN {\color{red}{underperforms}} Viterbi.}  
    \label{full4}
\end{figure*}

% \onecolumn
% \begin{landscape}
% \section{Full Qualitative Figures}
% \label{app:qualitative}
% \input{figures/qualtiative_best_LSTM}
% \end{landscape}
% \input{figures/qualtiative_best_Viterbi}

% uncomment to generate supplementary materials without main text
% \bibliographystyle{named}
% \bibliography{references}

%\section{Please add supplemental material as appendix here}
%
%Put anything that you might normally include after the references as an appendix here, {\it not in a separate supplementary file}. Upload your final camera-ready as a single pdf, including all appendices.

\end{document}